%% file: main.tex
\documentclass[11pt, a4paper, onecolumn, copyright, gr]{google}

\usepackage[authoryear, sort&compress, round]{natbib}
\uselogo{} 

\title{TabFM-Auto: Self-Evolving Pipelines for Tabular Foundation Models}

\correspondingauthor{deqing@google.com, weihaokong@google.com}

\renewcommand{\today}{2026-09-29}

\newif\ificlr
\iclrfalse
\let\origttfamily\ttfamily
\renewcommand{\ttfamily}{\origttfamily\hyphenchar\font=`\-}
\let\origunderscore\_
\renewcommand{\_}{\origunderscore\allowbreak}
\newcommand{\tabfm}{\mbox{\texttt{TabFM}}}
\newcommand{\tabfmp}{\mbox{\texttt{TabFM+}}}
\newcommand{\tabfmauto}{\mbox{\texttt{TabFM-Auto}}}
\definecolor{tabfmautored}{HTML}{D93025}
\definecolor{tabfmblue}{HTML}{3186FF}
\newcommand{\best}[1]{\textbf{#1}}
\usepackage[table,dvipsnames]{xcolor}
\usepackage{colortbl}
\usepackage{booktabs}
\usepackage{multirow}
\usepackage{enumitem}
\usepackage[most]{tcolorbox}  % 'most' loads 'raster' (tcbitemize) and 'skins' (enhanced)
\usepackage{listings}
\definecolor{tabfmautored}{HTML}{C5221F}  % Case I: Explicit Scientific Knowledge
\definecolor{tabfmblue}{HTML}{1967D2}     % Case II: Structural & Relational FE
\definecolor{tabfmteal}{HTML}{00796B}     % Case III: Model Calibration Only
\definecolor{codebg}{HTML}{F8FAFC}
\definecolor{codeframe}{HTML}{CBD5E1}
\definecolor{kwcolor}{HTML}{B91C1C}
\definecolor{strcolor}{HTML}{047857}
\definecolor{cmtcolor}{HTML}{64748B}
\lstdefinestyle{pybox}{
  language=Python,
  basicstyle=\ttfamily\scriptsize,
  keywordstyle=\bfseries\color{kwcolor},
  stringstyle=\color{strcolor},
  commentstyle=\itshape\color{cmtcolor},
  backgroundcolor=\color{codebg},
  frame=single,
  rulecolor=\color{codeframe},
  framesep=3pt,
  xleftmargin=3pt,
  xrightmargin=3pt,
  aboveskip=4pt,
  belowskip=4pt,
  showstringspaces=false,
  breaklines=true,
  columns=fullflexible
}

\author[1,3]{Deqing Fu}
\author[2,4]{Huangyuan Su}
\author[1]{Rajat Sen}
\author[1]{Taman Narayan}
\author[1,5]{Sujay Sanghavi}
\author[1]{Abhimanyu Das}
\author[1]{Weihao Kong}

\affil[1]{Google Research}
\affil[2]{Google DeepMind}
\affil[3]{University of Southern California}
\affil[4]{Harvard University}
\affil[5]{University of Texas at Austin}

\input{sections/abstract}

\begin{document}
\renewcommand{\topfraction}{0.99}
\renewcommand{\dbltopfraction}{0.99}
\renewcommand{\bottomfraction}{0.99}
\renewcommand{\textfraction}{0.01}
\renewcommand{\floatpagefraction}{0.85}
\renewcommand{\dblfloatpagefraction}{0.85}
\Urlmuskip=0mu plus 1mu\relax
\def\UrlBreaks{\do\/\do-\do\.}
\setlength{\emergencystretch}{2em}

\maketitle

\begin{figure*}[htp]
    \includegraphics[width=\linewidth]{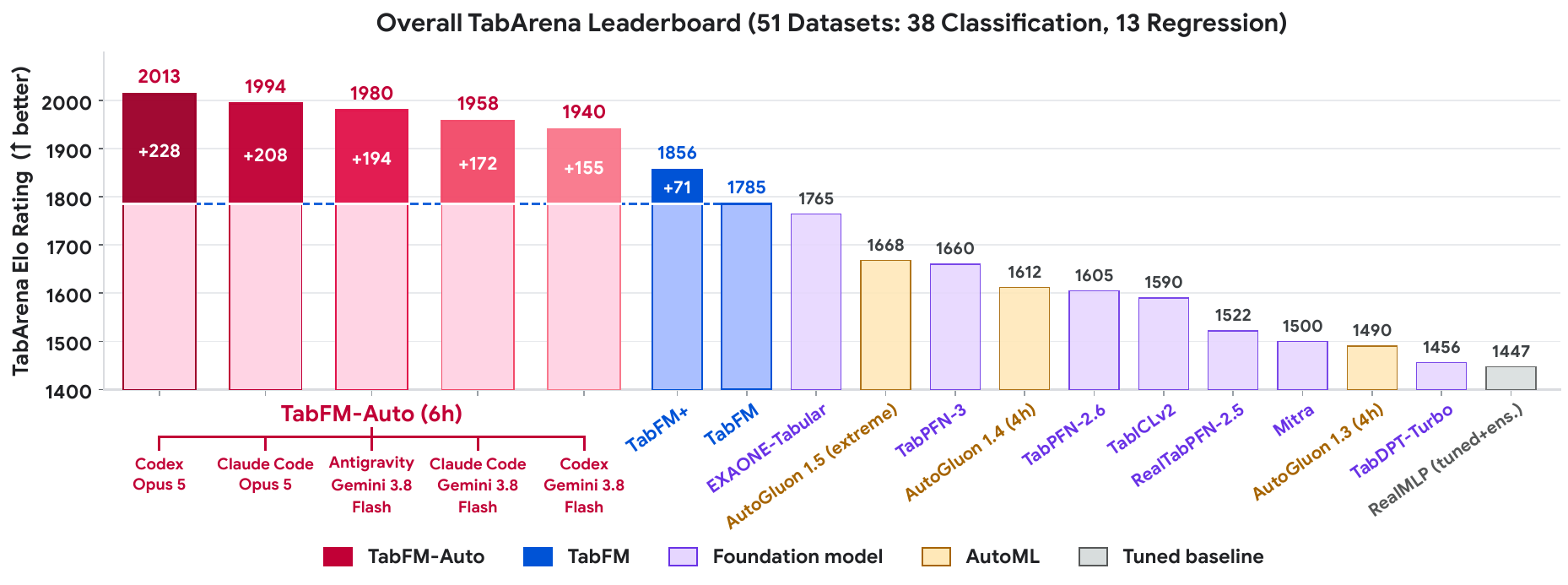}
    \caption{\textbf{Overall TabArena leaderboard across all 51 datasets (38 classification, 13 regression).} All five \tabfmauto{} configurations (red) take the top five positions, reaching up to 2013 Elo ($+228$ over the \tabfm{} baseline) and outperforming all other tabular foundation models.}
    \label{fig:headline}
\end{figure*}

\input{sections/01Introduction}
\input{sections/02RelatedWork}
\input{sections/03Method}
\input{sections/04Experiments}

\input{sections/05Conclusion}

{\frenchspacing
\bibliography{main}
}

\clearpage
\appendix
\input{sections/06Appendix}
\end{document}

%% file: sections/abstract.tex
\begin{abstract}
Tabular foundation models achieve strong zero-shot accuracy on structured data by pretraining on synthetic tables, but they ignore the column names, task descriptions, and auxiliary files that carry dataset semantics. Meanwhile, self-evolving machine learning engineering (MLE) agents train models from scratch on each dataset, yet jointly searching over features, architectures, and hyperparameters is noisy and prone to overfitting. We introduce \tabfmauto{}, which pairs a tabular foundation model, \tabfm{}, with a language model agent that evolves the data pipeline around it. Guided by dataset metadata and validation feedback, \tabfmauto{} iteratively refines data cleaning, feature engineering, context selection, and post-processing to reduce \tabfm{}'s error. Across all 51 datasets of the TabArena benchmark, five \tabfmauto{} configurations with different agents and language models take the top five overall positions, and the best raises \tabfm{} from 1785 to 2013 Elo. The discovered pipelines also transfer to other frozen tabular foundation models ($+69$ to $+143$ Elo) with no further search. On the 8 tabular competitions of MLE-Bench, \tabfmauto{} ranks first overall among MLE agents.
\end{abstract}

%% file: sections/01Introduction.tex
\section{Introduction} \label{sec:intro}

\begin{figure}[t]
    \centering
    \includegraphics[width=\linewidth]{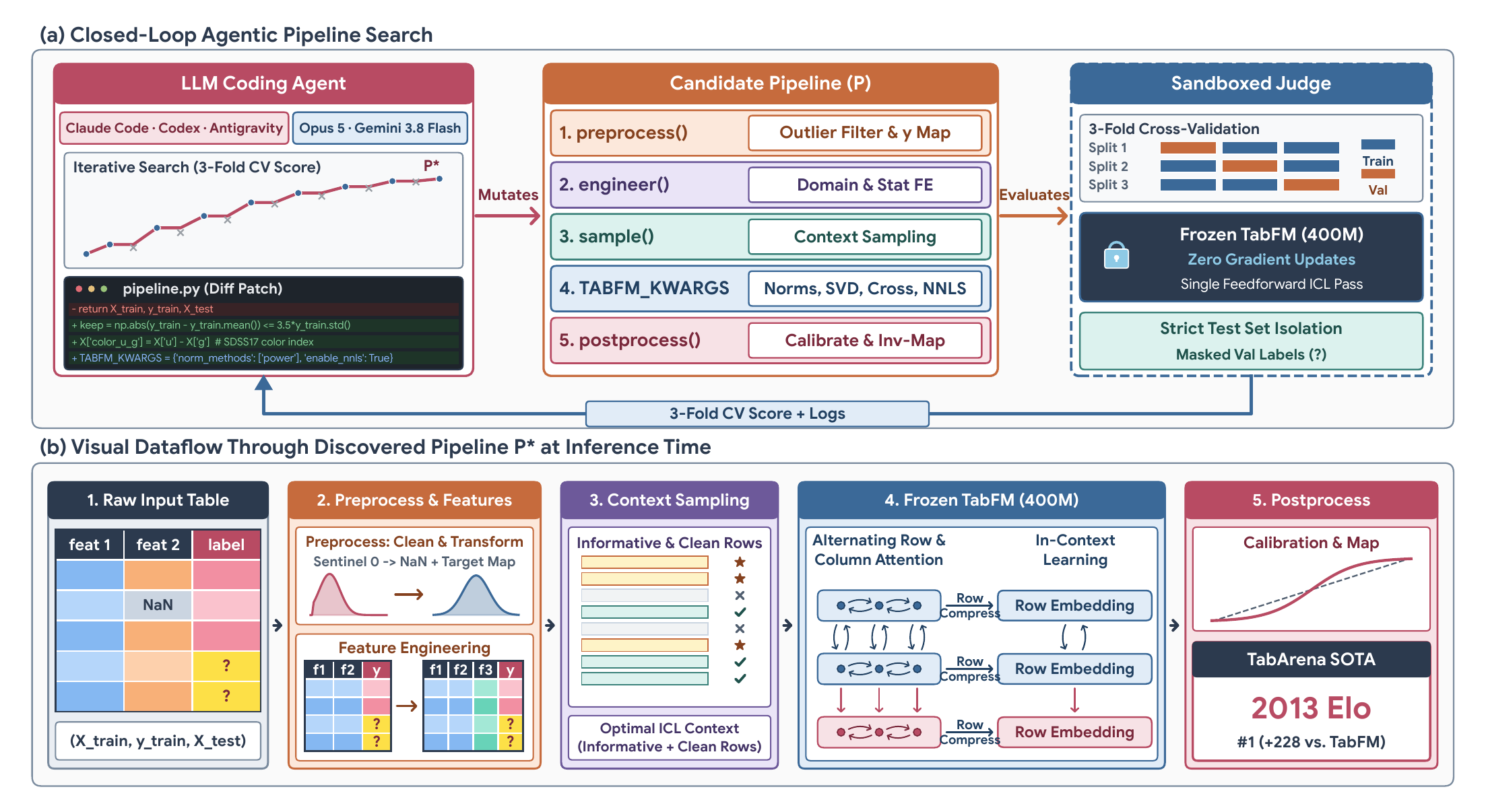}
    \caption{\textbf{Overview of \tabfmauto{}.} (a) Within a sandboxed evaluator, a coding agent repeatedly edits the data pipeline $P$ around a frozen 400M-parameter \tabfm{} model using 3-fold cross-validation on the training set. (b) At test time, the selected pipeline $P^*$ transforms the input table through data cleaning, feature engineering, context sampling, frozen \tabfm{} inference, and post-processing.}
    \label{fig:overview}
\end{figure}

Tabular Foundation Models (TFMs) such as TabPFN \citep{hollmann2023tabpfn, hollmann2025tabpfn2}, TabICL \citep{qu2025tabicl, qu2026tabiclv2}, and TabFM \citep{tabfm2026} pretrain a transformer on synthetic datasets so it can predict on new tables in a single forward pass without gradient updates \citep{muller2022pfn}. Although TFMs match or outperform tuned gradient-boosted decision trees (GBDTs) \citep{chen2016xgboost, ke2017lightgbm, prokhorenkova2018catboost}, their attention layers see only floating-point magnitudes and category indices. For example, they treat a column named \texttt{systolic\_bp} or \texttt{ICD9\_diagnosis} the same as an anonymous column \texttt{X\_17}. Thus, a TFM cannot infer domain formulas from column names, group clinical codes by hierarchy, or recognize when a number such as \texttt{0} encodes a missing measurement. Prior work (\tabfmp{}, \citealp{tabfm2026}) extends \tabfm{} with pairwise cross features, singular value decomposition (SVD) projections, and multi-view ensembling, but these generic operations ignore what the columns represent.

Pretrained on scientific text and code, Large Language Models (LLMs) understand the domain meaning of column names and task descriptions, yet they perform poorly as direct tabular predictors due to context-length limits, coarse number tokenization \citep{hegselmann2023tabllm, zhou2024pretrained, zhou2025fone, fu2026convergent}, and miscalibrated probabilities. Existing machine learning engineering (MLE) agents \citep{jiang2025aide, yang2025rdagent, du2026mlevolve} instead use LLMs to edit end-to-end training scripts that fit tree ensembles or neural networks from scratch during search. However, searching over features, model architectures, and hyperparameters at the same time is noisy: training variance often masks small feature gains and leads to validation overfitting.

We introduce \tabfmauto{}, which pairs an LLM agent with a frozen tabular foundation model, \tabfm{}. Guided by dataset metadata, the agent iteratively edits a data pipeline for data cleaning, feature engineering, context selection, and post-processing via 3-fold cross-validation on the training set while keeping \tabfm{} fixed. Since \tabfm{} requires no gradient training, each candidate pipeline is fast to evaluate and free of model-retraining noise. After search finishes, we evaluate the best pipeline once on the official test set. The pipeline also selects context rows on large or imbalanced tables and calibrates probabilities to skewed class priors.

We test five \tabfmauto{} configurations that combine three agent harnesses, \texttt{Claude Code} (\texttt{CC}), \texttt{Antigravity} (\texttt{AGY}), and \texttt{Codex}, with two language models (\texttt{Claude Opus 5} and \texttt{Gemini 3.8 Flash}). On all 51 datasets of the TabArena benchmark \citep{erickson2025tabarena}, they take the top five overall positions and outperform tuned AutoGluon ensembles and all evaluated tabular foundation models. The top configuration (\texttt{Codex} with \texttt{Opus 5}) raises \tabfm{} from 1785.3 to 2013.0 Elo. The discovered pipelines transfer without further search to other tabular foundation models (up to a $+143.3$ Elo improvement). On the 8 tabular competitions of MLE-Bench \citep{chan2024mlebench}, \tabfmauto{} ranks first overall ahead of existing MLE agents.

%% file: sections/02RelatedWork.tex
\input{tables/leaderboard_cls_reg_side_by_side}

\section{Related Work}
\label{sec:related}

\paragraph{Tabular Foundation Models and In-Context Optimization.}
Prior-Data Fitted Networks \citep{muller2022pfn} pretrain transformers on synthetic datasets so that a single forward pass approximates Bayesian in-context prediction \citep{garg2022what, vonoswald2023transformers, li2023transformers, fu2024second}. TabPFN \citep{hollmann2023tabpfn} introduced this paradigm for small classification tables. Later work extended it to regression and larger contexts \citep{hollmann2025tabpfn2, grinsztajn2025tabpfn25, grinsztajn2026tabpfn3}, added continued pretraining on real data \citep{ma2024tabdpt, garg2025realtabpfn, hosseinzadeh2026tabdptturbo}, and made attention linear-time with inducing points \citep{qu2025tabicl, qu2026tabiclv2}. Since these models are pretrained on anonymous synthetic variables, they typically ignore column names and task descriptions at test time. Even \tabfmp{} \citep{tabfm2026} only adds generic feature crosses, SVD projections, and view ensembling. Other work adds language encoders to tabular models to read column names or text cells \citep{kim2024carte, yan2024tpberta, tajjar2026tabpfntext}. Rather than modifying the tabular foundation model architecture, \tabfmauto{} keeps \tabfm{} frozen and uses an LLM agent to turn column names and task metadata into explicit pipeline code.

\paragraph{AutoML, LLM Feature Engineering, and MLE Agents.}
Supervised learning on tabular data has traditionally relied on GBDTs \citep{chen2016xgboost, ke2017lightgbm, prokhorenkova2018catboost} and deep tabular networks \citep{gorishniy2021revisiting, gorishniy2024tabm}. Classical automated machine learning (AutoML) systems such as Auto-sklearn \citep{feurer2015autosklearn} and AutoGluon-Tabular \citep{erickson2020autogluon} automate model selection and multi-layer stacking across these estimators. Prior LLM feature-engineering methods such as CAAFE \citep{hollmann2023caafe}, FeatLLM \citep{han2024featllm}, and OCTree \citep{nam2024octree} prompt an LLM to append derived columns to single tables. \tabfmauto{} extends this loop to the full data pipeline around an in-context model, including data cleaning, context-row selection, output calibration, and reading auxiliary files within a code-execution sandbox. Other systems automate end-to-end data science and multi-agent workflows \citep{huang2023mlagentbench, tornede2024automl, qi2026economy} or evolve programs through LLM-guided search \citep{romera2024funsearch, novikov2025alphaevolve, xu2026bes}, as \tabfmauto{} does for data pipelines. End-to-end agents such as AIDE \citep{jiang2025aide}, R\&D-Agent \citep{yang2025rdagent}, and MLEvolve \citep{du2026mlevolve} retrain tree ensembles or neural networks during search, whereas \tabfmauto{} keeps \tabfm{} frozen and searches only over the data pipeline.

%% file: tables/leaderboard_cls_reg_side_by_side.tex
% Generated by tabfm_auto_paper/scripts/generate_winrates_and_tables.py
\begin{table*}[t]
\centering
\caption{\textbf{Side-by-side TabArena Classification (38 datasets) and Regression (13 datasets) leaderboards.} Each \tabfmauto{} configuration is evaluated individually against the 66-method TabArena pool~\citep{erickson2025tabarena}: \textbf{Elo} measures pairwise rating, \textbf{Wins} counts datasets where a method ranks first, \textbf{Improv.} measures mean relative error gap to the per-dataset best model, and \textbf{G-Mean} is the geometric mean test error across datasets.}
\label{tab:cls_reg_sbs}
\label{tab:cls}
\label{tab:reg}
\setlength{\tabcolsep}{4.0pt}
\begin{minipage}[t]{0.48\textwidth}
\centering
\small\textbf{(a) Classification (38 Datasets)}\\[3pt]
\resizebox{\linewidth}{!}{%
\begin{tabular}{clcccc}
\toprule
\textbf{\#} & \textbf{Method} & \textbf{Elo} $\uparrow$ & \textbf{Wins} $\uparrow$ & \textbf{Improv.} $\downarrow$ & \textbf{G-Mean} $\downarrow$\\
\midrule
\rowcolor{tabfmautored!12}
1 & TabFM-Auto (Codex, Opus 5) & \best{1966.3} & 13.84 & 1.91\% & 0.0913\\
\rowcolor{tabfmautored!12}
2 & TabFM-Auto (AGY, Gemini 3.8 Flash) & 1937.5 & \best{16.11} & 2.45\% & \best{0.0902}\\
\rowcolor{tabfmautored!12}
3 & TabFM-Auto (CC, Opus 5) & 1933.9 & 14.93 & \best{1.69\%} & 0.0913\\
\rowcolor{tabfmautored!12}
4 & TabFM-Auto (CC, Gemini 3.8 Flash) & 1918.7 & 13.26 & 2.52\% & 0.0928\\
\rowcolor{tabfmautored!12}
5 & TabFM-Auto (Codex, Gemini 3.8 Flash) & 1875.0 & 15.44 & 2.43\% & 0.0924\\
6 & TabFM+ & 1836.7 & 12.22 & 2.91\% & 0.0948\\
7 & TabFM & 1768.7 & 7.91 & 3.80\% & 0.0957\\
8 & EXAONE-Tabular & 1759.5 & 4.39 & 7.28\% & 0.1009\\
9 & AutoGluon 1.5 (extreme) & 1664.6 & 1.72 & 7.86\% & 0.1010\\
10 & TabPFN-3 & 1637.7 & 0.87 & 10.18\% & 0.1050\\
11 & AutoGluon 1.4 (4h) & 1617.1 & 0.51 & 11.24\% & 0.1079\\
12 & TabPFN-2.6 & 1586.9 & 0.14 & 11.72\% & 0.1075\\
13 & TabICLv2 & 1583.6 & 1.08 & 10.88\% & 0.1060\\
14 & RealTabPFN-2.5 & 1533.5 & 0.07 & 12.11\% & 0.1077\\
15 & AutoGluon 1.3 (4h) & 1473.4 & 0.14 & 14.24\% & 0.1147\\
16 & RealMLP (tuned+ens.) & 1435.2 & 0.19 & 15.37\% & 0.1161\\
\bottomrule
\end{tabular}%
}
\end{minipage}%
\hfill
\begin{minipage}[t]{0.48\textwidth}
\centering
\small\textbf{(b) Regression (13 Datasets)}\\[3pt]
\resizebox{\linewidth}{!}{%
\begin{tabular}{clcccc}
\toprule
\textbf{\#} & \textbf{Method} & \textbf{Elo} $\uparrow$ & \textbf{Wins} $\uparrow$ & \textbf{Improv.} $\downarrow$ & \textbf{G-Mean} $\downarrow$\\
\midrule
\rowcolor{tabfmautored!12}
1 & TabFM-Auto (Codex, Opus 5) & \best{2512.9} & \best{11.12} & \best{0.00\%} & \best{15.86}\\
\rowcolor{tabfmautored!12}
2 & TabFM-Auto (CC, Opus 5) & 2489.7 & 10.51 & 0.00\% & 15.89\\
\rowcolor{tabfmautored!12}
3 & TabFM-Auto (Codex, Gemini 3.8 Flash) & 2423.2 & 9.58 & 0.01\% & 15.93\\
\rowcolor{tabfmautored!12}
4 & TabFM-Auto (CC, Gemini 3.8 Flash) & 2407.9 & 9.41 & 0.01\% & 15.95\\
\rowcolor{tabfmautored!12}
5 & TabFM-Auto (AGY, Gemini 3.8 Flash) & 2347.8 & 9.98 & 0.00\% & 15.88\\
6 & TabFM+ & 2169.2 & 5.33 & 0.14\% & 16.17\\
7 & TabFM & 2045.9 & 1.80 & 1.43\% & 16.40\\
8 & EXAONE-Tabular & 1967.8 & 1.17 & 2.46\% & 16.58\\
9 & TabPFN-3 & 1863.1 & 0.71 & 2.15\% & 16.51\\
10 & AutoGluon 1.5 (extreme) & 1851.2 & 0.39 & 3.67\% & 16.78\\
11 & TabPFN-2.6 & 1790.1 & 0.13 & 3.74\% & 16.79\\
12 & AutoGluon 1.4 (4h) & 1731.0 & 0.07 & 4.42\% & 16.91\\
13 & TabICLv2 & 1726.2 & 0.81 & 3.68\% & 16.78\\
14 & TabDPT-Turbo & 1661.9 & 0.26 & 4.93\% & 17.02\\
15 & AutoGluon 1.3 (4h) & 1646.1 & 0.03 & 6.01\% & 17.23\\
16 & RealMLP (tuned+ens.) & 1622.8 & 0.07 & 5.41\% & 17.09\\
\bottomrule
\end{tabular}%
}
\end{minipage}
\end{table*}

%% file: sections/03Method.tex
\section{Methodology}
\label{sec:method}
\subsection{Problem Formulation}
\label{sec:method_formulation}
Consider a tabular prediction task with a labeled training dataset $\mathcal{D}_{\text{train}} = (\mathbf{X}_{\text{train}}, \mathbf{y}_{\text{train}})$ of $T_{\text{train}}$ rows and $H$ columns, an unlabeled test dataset $\mathbf{X}_{\text{test}}$, and dataset metadata $\mathcal{M}$ containing column names, units, task descriptions, and auxiliary file schemas. A zero-shot tabular foundation model $f_{\theta^*}$ such as \tabfm{}~\citep{tabfm2026} takes $(\mathbf{X}_{\text{train}}, \mathbf{y}_{\text{train}})$ as in-context examples and predicts test labels $\hat{\mathbf{y}} = f_{\theta^*}(\mathbf{X}_{\text{test}} \mid \mathbf{X}_{\text{train}}, \mathbf{y}_{\text{train}})$ in a single forward pass with frozen weights $\theta^*$.

\paragraph{Optimization Objective.}
We partition $\mathcal{D}_{\text{train}}$ into 3 internal cross-validation folds without touching the held-out test split (Figure~\ref{fig:overview}). On each internal fold $(\mathcal{D}_{\text{subtrain}}, \mathcal{D}_{\text{val}})$ and conditioned on $\mathcal{M}$, the coding agent searches over executable pipelines $P = (\Phi_{\text{clean}}, \Phi_{\text{feat}}, \mathcal{S}_{\text{ctx}}, \Psi_{\text{post}}) \in \mathcal{P}$, defined below, to maximize the mean 3-fold cross-validation metric $\mathcal{U}_{\text{val}}$ around the fixed weights $\theta^*$:
\begin{equation}
\begin{aligned}
    P^* &= \arg\max_{P \in \mathcal{P}} \; \mathcal{U}_{\text{val}}\!\Big(\, \Psi_{\text{post}}\!\big(\, f_{\theta^*}\!\big(\tilde{\mathbf{X}}_{\text{val}} \;\big|\; \mathcal{S}_{\text{ctx}}(\tilde{\mathbf{X}}_{\text{subtrain}}, \tilde{\mathbf{y}}_{\text{subtrain}})\big),\; \mathbf{y}_{\text{subtrain}},\; \tilde{\mathbf{X}}_{\text{val}}\,\big),\; \mathbf{y}_{\text{val}} \,\Big), \\
    \text{where}\quad &(\tilde{\mathbf{X}}_{\text{subtrain}}, \tilde{\mathbf{y}}_{\text{subtrain}}, \tilde{\mathbf{X}}_{\text{val}}) = \Phi_{\text{feat}}\big(\Phi_{\text{clean}}(\mathbf{X}_{\text{subtrain}}, \mathbf{y}_{\text{subtrain}}, \mathbf{X}_{\text{val}})\big).
\end{aligned}
\label{eq:auto_obj}
\end{equation}
Throughout search, $\theta^*$ stays fixed, and passing unlabeled inputs $\mathbf{X}_{\text{val}}$ (or $\mathbf{X}_{\text{test}}$) with the training split while withholding evaluation labels lets the agent perform both \textit{inductive} and \textit{transductive} learning.

\subsection{Building Pipelines Around Frozen \tabfm{}}
\label{sec:method_hooks}

Each candidate pipeline $P$ implements the four stages in Equation~\eqref{eq:auto_obj} through four modular Python functions (\texttt{preprocess()}, \texttt{engineer()}, \texttt{sample()}, and \texttt{postprocess()}), together with the \texttt{TABFM\_KWARGS} configuration dictionary passed to $f_{\theta^*}$ (Figure~\ref{fig:overview}). Rather than rewriting an end-to-end training script, the agent edits only these components, and each stage addresses a specific limitation of a frozen TFM. The first two stages shape the table that \tabfm{} reads. \textbf{Data cleaning and target conditioning} ($\Phi_{\text{clean}}$) handles values whose meaning a TFM cannot read from magnitudes alone, such as a \texttt{0} that encodes a missing measurement. Using $\mathcal{M}$, \texttt{preprocess()} can recode such placeholders into missingness indicators, filter corrupted training rows, and apply reversible target transformations $g(\mathbf{y}_{\text{train}})$, such as $\log(1+y)$, that compress skewed targets. \textbf{Semantic feature engineering} ($\Phi_{\text{feat}}$) is where the LLM's domain knowledge enters the pipeline. Given the cleaned tables, \texttt{engineer()} translates $\mathcal{M}$ into explicit columns for both inductive features (domain ratios and formulas, group statistics, and auxiliary-file summaries) and label-free transductive features over combined train and test inputs (e.g., entity-graph degrees in Figure~\ref{fig:fe_case_studies}). Writing a formula as one column spares \tabfm{} from inferring it in context, which helps most on real-world schemas.

The remaining two stages control which rows \tabfm{} conditions on and how its outputs are used. \textbf{Context selection} ($\mathcal{S}_{\text{ctx}}$) is needed because \tabfm{} is pretrained on tables of up to 16,384 rows~\citep{tabfm2026} and could potentially degrade in performance on much larger tables, yet uniform subsampling can drop rare classes. \texttt{sample()} therefore builds multiple context subsets (views) that stratify by class or oversample minority classes, and combining predictions across views lets \tabfm{} use more rows than one context window holds. \textbf{Post-processing and output calibration} ($\Psi_{\text{post}}$) applies $g^{-1}$ to return regression predictions to the original target scale, which keeps candidates with different target transformations comparable. For classification, it can shift log-odds toward the training class prior or apply temperature or Platt scaling~\citep{platt1999probabilistic}, since oversampled views and synthetic pretraining data may not match the class balance and confidence of real data.

Generic versions of all four stages already exist in \tabfmp{}~\citep{tabfm2026}, which extends \tabfm{} to normalize inputs and clip outliers ($\Phi_{\text{clean}}$), append pairwise feature crosses and SVD projections ($\Phi_{\text{feat}}$), ensemble predictions over multiple views with a configurable context size ($\mathcal{S}_{\text{ctx}}$), and weight the views by non-negative least squares (NNLS)~\citep{lawson1995solving} before calibrating the output ($\Psi_{\text{post}}$). \textbf{\texttt{TABFM\_KWARGS}} exposes these overlapping operations as constructor settings of $f_{\theta^*}$. The agent can thus cover the generic parts of a stage by tuning settings instead of writing code, and use the four functions for what the \tabfmp{} presets cannot express, such as domain formulas or class-balanced contexts. Keeping these settings apart from the four functions also lets us transfer the agent's searched pipelines unchanged around other TFMs (\S\ref{sec:exp_tabarena}).

The agent self-evolves $P$ by \textbf{iterative search}. Every run starts from the identity pipeline $P_0$ (Appendix Figure~\ref{fig:identity_pipeline}), which feeds the unmodified table to \tabfm{} with default settings, and the agent keeps evolving $P$ to improve the cross-validation score $\mathcal{U}_{\text{val}}$. Because $\theta^*$ stays frozen, candidates need no gradient training, no retraining noise masks small feature gains, and every improvement over $P_0$ comes from the agent's new pipeline. To hide test labels from agent-written code, the evaluation harness runs every candidate in a \textbf{sandbox} without network access or the held-out test splits (Figure~\ref{fig:overview}a and Appendix~\ref{sec:app_verification}). After search, the harness freezes $P^*$ and scores the official test splits.

\input{figures/tabarena_case_study}

%% file: figures/tabarena_case_study.tex
\begin{figure}[t]
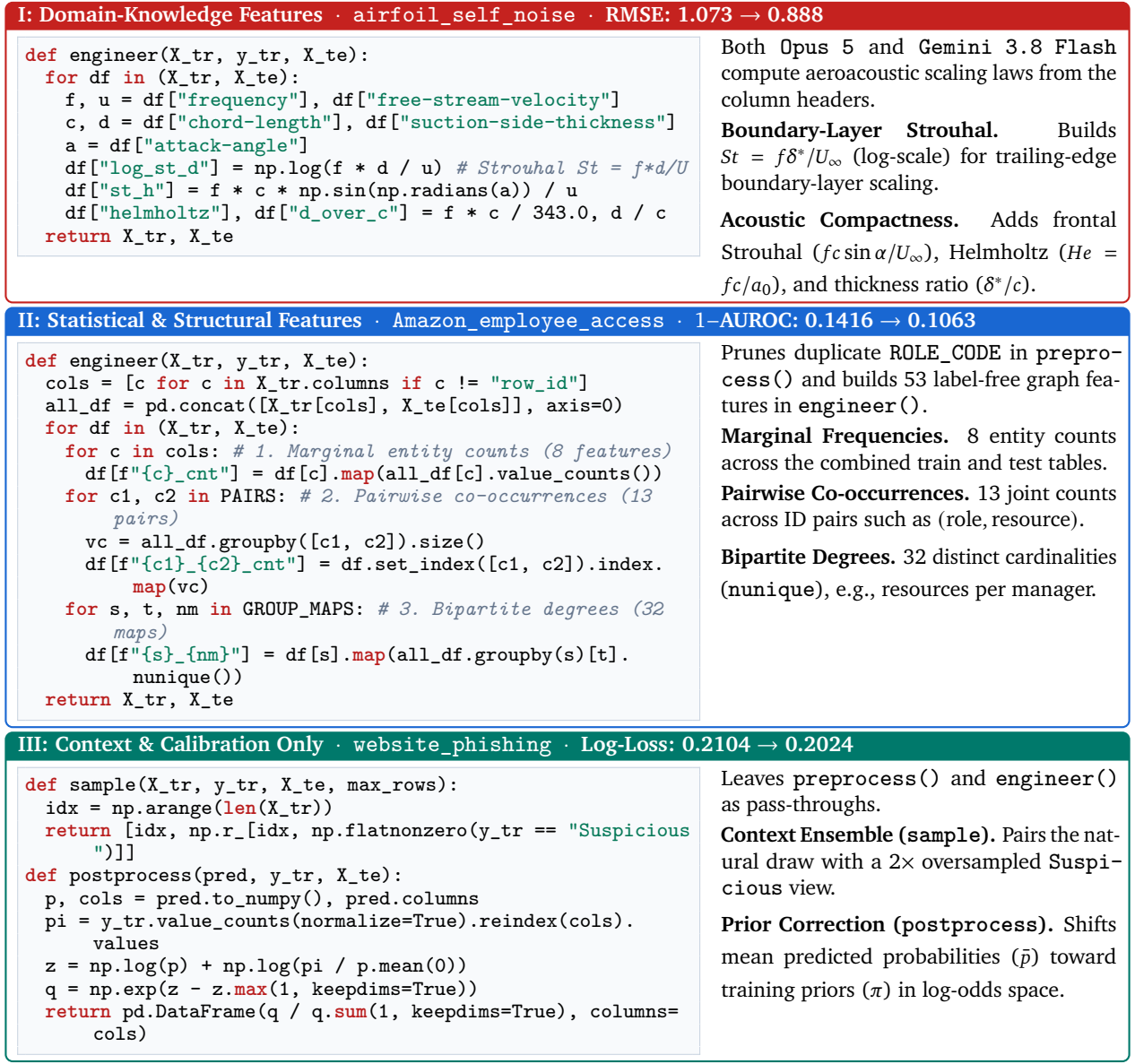

\centering
\begin{tcbitemize}[
  raster columns=1,
  raster row skip=1.5pt,
  enhanced,
  colback=white,
  fonttitle=\bfseries\footnotesize,
  boxrule=0.8pt,
  arc=2.5pt,
  boxsep=0.8pt,
  left=4pt, right=4pt, top=2.5pt, bottom=1.5pt
]
% ==================== CARD I ====================
\tcbitem[
  colframe=tabfmautored,
  colbacktitle=tabfmautored,
  coltitle=white,
  title={I: Domain-Knowledge Features $\;\cdot\;$ \normalfont\footnotesize\color{white}\texttt{airfoil\_self\_noise} $\;\cdot\;$ \textbf{RMSE:} \textbf{1.073 $\to$ 0.888}}
]
\begin{minipage}[t]{0.62\textwidth}
\vspace{0pt}
\begin{lstlisting}[style=pybox,basicstyle=\ttfamily\scriptsize,aboveskip=4.5pt,belowskip=0pt]
def engineer(X_tr, y_tr, X_te):
  for df in (X_tr, X_te):
    f, u = df["frequency"], df["free-stream-velocity"]
    c, d = df["chord-length"], df["suction-side-thickness"]
    a = df["attack-angle"]
    df["log_st_d"] = np.log(f * d / u)  # Strouhal St = f*d/U
    df["st_h"] = f * c * np.sin(np.radians(a)) / u
    df["helmholtz"], df["d_over_c"] = f * c / 343.0, d / c
  return X_tr, X_te
\end{lstlisting}
\end{minipage}%
\hfill
\begin{minipage}[t]{0.36\textwidth}
\vspace{0pt}
{\footnotesize\setlength{\parskip}{2pt}\setlength{\parindent}{0pt}%
Both \texttt{Opus~5} and \texttt{Gemini~3.8 Flash} compute aeroacoustic scaling laws from the column headers.\par
\textbf{Boundary-Layer Strouhal.} Builds $St = f\delta^*/U_\infty$ ($\log$-scale) for trailing-edge boundary-layer scaling.\par
\textbf{Acoustic Compactness.} Adds frontal Strouhal ($f c \sin\alpha / U_\infty$), Helmholtz ($He = f c / a_0$), and thickness ratio ($\delta^*/c$).}
\end{minipage}

% ==================== CARD II ====================
\tcbitem[
  colframe=tabfmblue,
  colbacktitle=tabfmblue,
  coltitle=white,
  title={II: Statistical \& Structural Features $\;\cdot\;$ \normalfont\footnotesize\color{white}\texttt{Amazon\_employee\_access} $\;\cdot\;$ \textbf{$1-$AUROC:} \textbf{0.1416 $\to$ 0.1063}}
]
\begin{minipage}[t]{0.62\textwidth}
\vspace{0pt}
\begin{lstlisting}[style=pybox,basicstyle=\ttfamily\scriptsize,aboveskip=4.5pt,belowskip=0pt]
def engineer(X_tr, y_tr, X_te):
  cols = [c for c in X_tr.columns if c != "row_id"]
  all_df = pd.concat([X_tr[cols], X_te[cols]], axis=0)
  for df in (X_tr, X_te):
    for c in cols:  # 1. Marginal entity counts (8 features)
      df[f"{c}_cnt"] = df[c].map(all_df[c].value_counts())
    for c1, c2 in PAIRS:  # 2. Pairwise co-occurrences (13 pairs)
      vc = all_df.groupby([c1, c2]).size()
      df[f"{c1}_{c2}_cnt"] = df.set_index([c1, c2]).index.map(vc)
    for s, t, nm in GROUP_MAPS:  # 3. Bipartite degrees (32 maps)
      df[f"{s}_{nm}"] = df[s].map(all_df.groupby(s)[t].nunique())
  return X_tr, X_te
\end{lstlisting}
\end{minipage}%
\hfill
\begin{minipage}[t]{0.36\textwidth}
\vspace{0pt}
{\footnotesize\setlength{\parskip}{2pt}\setlength{\parindent}{0pt}%
Prunes duplicate \texttt{ROLE\_CODE} in \texttt{preprocess()} and builds 53 label-free graph features in \texttt{engineer()}.\par
\textbf{Marginal Frequencies.} 8 entity counts across the combined train and test tables.\par
\textbf{Pairwise Co-occurrences.} 13 joint counts across ID pairs such as $(\text{role}, \text{resource})$.\par
\textbf{Bipartite Degrees.} 32 distinct cardinalities (\texttt{nunique}), e.g., resources per manager.}
\end{minipage}

% ==================== CARD III ====================
\tcbitem[
  colframe=tabfmteal,
  colbacktitle=tabfmteal,
  coltitle=white,
  title={III: Context \& Calibration Only $\;\cdot\;$ \normalfont\footnotesize\color{white}\texttt{website\_phishing} $\;\cdot\;$ \textbf{Log-Loss:} \textbf{0.2104 $\to$ 0.2024}}
]
\begin{minipage}[t]{0.62\textwidth}
\vspace{0pt}
\begin{lstlisting}[style=pybox,basicstyle=\ttfamily\scriptsize,aboveskip=4.5pt,belowskip=0pt]
def sample(X_tr, y_tr, X_te, max_rows):
  idx = np.arange(len(X_tr))
  return [idx, np.r_[idx, np.flatnonzero(y_tr == "Suspicious")]]
def postprocess(pred, y_tr, X_te):
  p, cols = pred.to_numpy(), pred.columns
  pi = y_tr.value_counts(normalize=True).reindex(cols).values
  z = np.log(p) + np.log(pi / p.mean(0))
  q = np.exp(z - z.max(1, keepdims=True))
  return pd.DataFrame(q / q.sum(1, keepdims=True), columns=cols)
\end{lstlisting}
\end{minipage}%
\hfill
\begin{minipage}[t]{0.36\textwidth}
\vspace{0pt}
{\footnotesize\setlength{\parskip}{2pt}\setlength{\parindent}{0pt}%
Leaves \texttt{preprocess()} and \texttt{engineer()} as pass-throughs.\par
\textbf{Context Ensemble (\texttt{sample}).} Pairs the natural draw with a $2\times$ oversampled \texttt{Suspicious} view.\par
\textbf{Prior Correction (\texttt{postprocess}).} Shifts mean predicted probabilities ($\bar{p}$) toward training priors ($\pi$) in log-odds space.}
\end{minipage}
\end{tcbitemize}
\caption{\textbf{Representative \tabfmauto{} pipelines across the three categories on TabArena (\texttt{Antigravity} with \texttt{Gemini 3.8 Flash}).} (I) Explicit aeroacoustic scaling laws on \texttt{airfoil\_self\_noise}, (II) relational graph features on \texttt{Amazon\_employee\_access}, and (III) minority-class context ensembling and prior calibration on \texttt{website\_phishing}.}
\label{fig:fe_case_studies}
\end{figure}

%% file: sections/04Experiments.tex
\section{Experiments}
\label{sec:expt}

We evaluate \tabfmauto{} on two benchmarks: against tabular foundation models and tuned AutoML ensembles on the 51-dataset TabArena benchmark (\S\ref{sec:exp_tabarena}), and against MLE agents that train models from scratch on MLE-Bench-Tabular (\S\ref{sec:exp_mlebench}).

\subsection{TabArena}
\label{sec:exp_tabarena}

\paragraph{Experimental Setup.}
We evaluate \tabfmauto{} on all 51 datasets of the TabArena benchmark~\citep{erickson2025tabarena} (38 classification and 13 regression), where each dataset uses 3-fold outer cross-validation repeated 3 or 10 times, yielding 9 or 30 official train/test splits (evaluation folds in Table~\ref{tab:all_51_datasets}). Running a separate LLM agent search on every fold would be prohibitively costly in tokens and GPU time (Table~\ref{tab:token_consumption}). Thus, we run pipeline search once per dataset using 3-fold cross-validation on the training set of fold~0 (up to 96 evaluations or 6 hours on one H100 GPU), allowing both inductive and transductive features while withholding evaluation labels. We then freeze $P^*$, evaluate it on the held-out test splits of all folds $k \in \{0, \dotsc, S-1\}$, and report both all-folds and fold-0-only test leaderboards (Tables~\ref{tab:overall}--\ref{tab:fold0_elo}). We test five configurations combining three agent harnesses (\texttt{Claude Code}, \texttt{Codex}, and \texttt{Antigravity}) with two language models (\texttt{Claude Opus~5} and \texttt{Gemini~3.8 Flash}). We compare against \tabfm{}, \tabfmp{}~\citep{tabfm2026}, other tabular foundation models (\texttt{EXAONE-Tabular}, \texttt{TabPFN-3}, \texttt{TabICLv2}), and 4-hour \texttt{AutoGluon} ensembles~\citep{erickson2020autogluon}. Following \citet{erickson2025tabarena}, we rate each configuration individually against the 66-method TabArena pool using Bradley--Terry \textbf{Elo}~\citep{bradley1952rank, elo1978rating, hunter2004mm} (anchored to default Random Forest $= 1000$), dataset \textbf{Wins}, oracle \textbf{Improvability} ($1 - \mathrm{err}_{\text{best}}/\mathrm{err}_{\text{method}}$), and Geometric Mean Test Error (\textbf{G-Mean}).

\paragraph{Main Results on TabArena.}
Across all 51 datasets (Figure~\ref{fig:headline}, Appendix~\ref{sec:app_leaderboard_analysis}, and Appendix~\ref{sec:app_per_dataset} Table~\ref{tab:all_51_datasets}), our five \tabfmauto{} configurations take the top five overall positions, reaching 2013.0 Elo (\texttt{Codex} with \texttt{Opus 5}) and outperforming 4-hour \texttt{AutoGluon 1.5 (extreme)} (1668.4 Elo) by $+272$ to $+345$ Elo. The gains hold on both task types (Table~\ref{tab:cls_reg_sbs} and Appendix Figure~\ref{fig:leaderboards}). On the 38 classification datasets, \tabfmauto{} raises \tabfm{} by $+197.6$ Elo to 1966.3 Elo, and \texttt{Antigravity} with \texttt{Gemini 3.8 Flash} has the lowest G-Mean test error (0.0902) and the most dataset wins (16.11). On the 13 regression datasets, \tabfmauto{} raises \tabfm{} by $+467.0$ Elo to 2512.9 Elo, improving over \tabfm{} on all 13 datasets and reducing oracle improvability to $0.00\%$. We hypothesize that regression gains are larger because linearizing physical ratios in $\Phi_{\text{feat}}$ and skewed targets in $\Phi_{\text{clean}}$ lets \tabfm{} smoothly interpolate continuous functions that tree splits approximate only coarsely.

Pairwise fold win rates on the official test sets (Figure~\ref{fig:tabarena_win_rate}) agree with the Elo ranking. \texttt{Claude Code} with \texttt{Opus 5} and \texttt{Antigravity} with \texttt{Gemini 3.8 Flash} tie head-to-head ($50.7\%$ vs.\ $49.3\%$). They win against \tabfm{} on $74.3\%$ and $72.1\%$ of test folds ($91.9\%$ and $83.3\%$ on regression) and against \texttt{AutoGluon 1.5 (extreme)} on $91.3\%$ and $89.5\%$. Even the fifth-ranked configuration (\texttt{Codex} with \texttt{Gemini 3.8 Flash}) reaches 1940.1 Elo.

\begin{figure}[t]
    \centering
    \includegraphics[width=\linewidth]{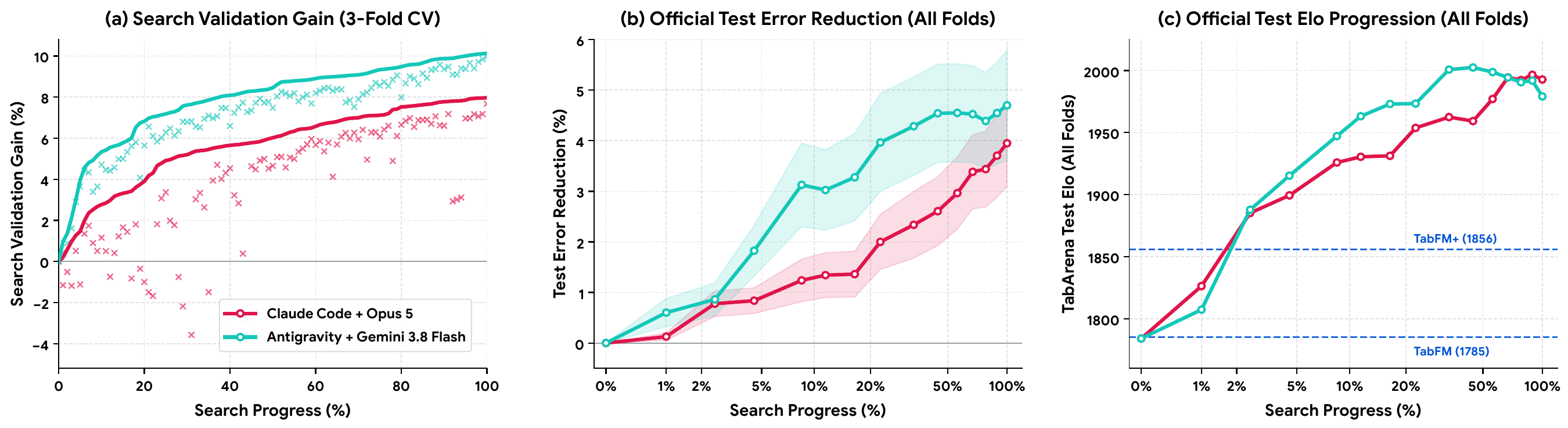}
    \caption{\textbf{Search dynamics and test generalization.} (a) 3-fold cross-validation gain on fold~0's training set, (b) official test error reduction across all folds, and (c) official test TabArena Elo progression across all folds.}
    \label{fig:hill_climbing}
\end{figure}

\paragraph{Domain Knowledge vs.\ Statistical Feature Engineering.}
To understand the strategies the agents discover, we inspect the final pipelines from \texttt{Claude Code} with \texttt{Opus 5} and \texttt{Antigravity} with \texttt{Gemini 3.8 Flash} for all 51 TabArena datasets (Figure~\ref{fig:fe_case_studies} and Appendix~\ref{sec:app_taxonomy_analysis} Table~\ref{tab:feature_engineering_taxonomy}). We group them into three categories: domain-knowledge features, statistical and structural features, and context and calibration changes only. The agents modify the feature table in $95.1\%$ of runs ($97/102$ across both configurations). They pair feature engineering ($\Phi_{\text{feat}}$) with missing-value cleaning and target transforms ($\Phi_{\text{clean}}$) on regression ($3.11\%\text{--}3.15\%$ lower G-Mean root mean squared error, RMSE), and with minority-class context selection ($\mathcal{S}_{\text{ctx}}$) and prior calibration ($\Psi_{\text{post}}$) on multiclass classification ($7.64\%\text{--}8.39\%$ lower G-Mean log-loss). On the 17 datasets whose column names or task descriptions identify real-world quantities, such as physical, clinical, or economic variables (Cat.~I), both agents write domain formulas directly from the schema (e.g., aerodynamic Strouhal numbers on \texttt{airfoil\_self\_noise}, $14.6\%\text{--}17.3\%$ lower test RMSE), reducing official test error across all folds by $7.3\%$ (\texttt{Opus 5}) to $8.2\%$ (\texttt{Gemini 3.8 Flash}). That is nearly three times the $2.3\%\text{--}3.0\%$ error reduction on the 34 datasets for \texttt{Opus 5} (29 for \texttt{Gemini 3.8 Flash}) with anonymized or generic schemas (Cat.~II), where the agent relies on statistical transforms, frequency encodings, and graph degree features. Cat.~I datasets account for 6 of the 10 largest test error reductions of each configuration. On the 5 runs where the agent does not change the feature table (Cat.~III), 4 runs improve through context sampling in $\mathcal{S}_{\text{ctx}}$ and/or prior calibration in $\Psi_{\text{post}}$, and the single run that tunes only hyperparameters is $0.30\%$ worse.

\paragraph{Validation-to-Test Generalization.}
To test whether iterative search overfits the 3-fold cross-validation split of fold~0, we evaluate up to 15 intermediate pipeline checkpoints per run from \texttt{Claude Code} with \texttt{Opus 5} and \texttt{Antigravity} with \texttt{Gemini 3.8 Flash}, taken at fixed fractions of the search budget, on the official test sets of all folds (Figure~\ref{fig:hill_climbing} and Appendix~\ref{sec:app_dynamics} Figures~\ref{fig:hill_climbing_split}--\ref{fig:overfitting_symlog}). We find that cross-validation gains on the training set of fold~0 (Figure~\ref{fig:hill_climbing}a) carry over to the official test sets: mean test error falls by $4.0\%$ (\texttt{Opus 5}) to $4.7\%$ (\texttt{Gemini 3.8 Flash}) from $P_0$ to full budget (Figure~\ref{fig:hill_climbing}b), and G-Mean error by $4.19\%$ and $5.05\%$ (Table~\ref{tab:all_51_datasets}). Both configurations score higher than \tabfmp{} (1856 Elo) within the first $2.5\%$ of the search (about 3 evaluations) and reach 1979--1993 Elo at full budget (Figure~\ref{fig:hill_climbing}c). On the held-out test split of fold~0 alone (Table~\ref{tab:fold0_elo}), \tabfmauto{} also holds the \#1 Overall, Classification, and Regression Elo ratings (Appendix~\ref{sec:app_leaderboard_analysis}).

\input{tables/free_agent_ablation}
\begin{figure}[t]
    \centering
    \includegraphics[width=\linewidth]{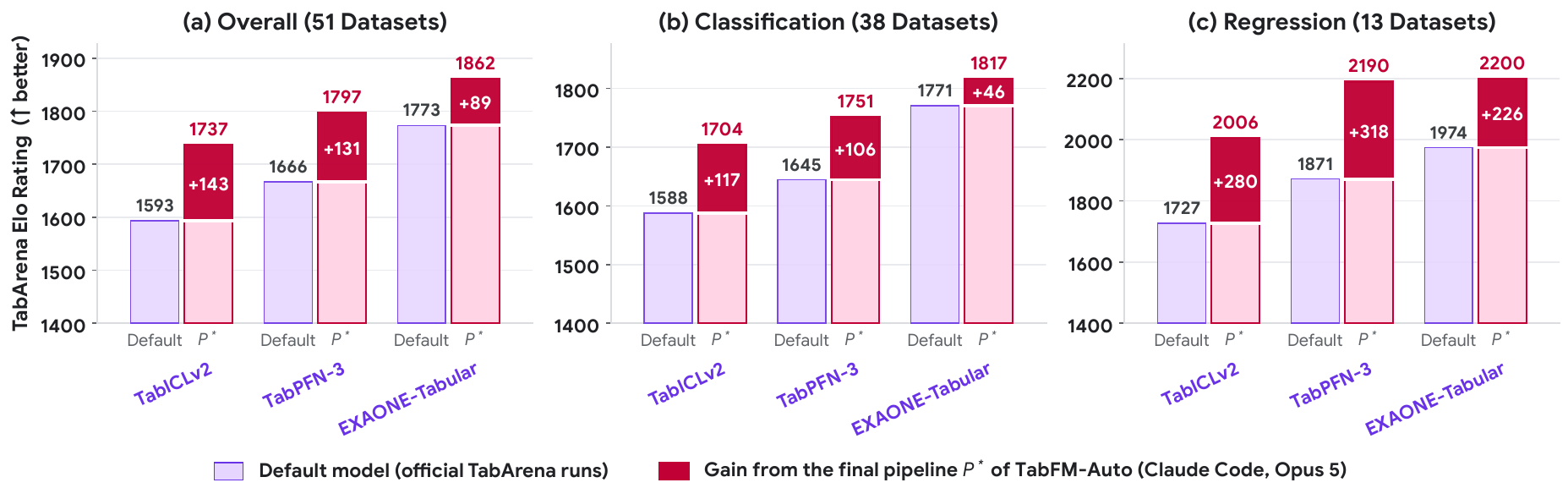}
    \caption{\textbf{Final pipelines found for \tabfm{} also help other tabular foundation models.} \emph{Default} is each model's official TabArena result, rated in the same fit as $P^*$. $P^*$ runs the same frozen model within the final per-dataset pipelines that \tabfmauto{} (\texttt{Claude Code}, \texttt{Opus 5}) found for \tabfm{}, keeping only the context size from \texttt{TABFM\_KWARGS} and without further search.}
    \label{fig:transfer}
\end{figure}

\paragraph{Ablation: Evolving Around Frozen \tabfm{} vs.\ an Unconstrained MLE Agent.}
We compare \tabfmauto{} against an unconstrained coding agent that uses the same \texttt{Antigravity} harness, \texttt{Gemini 3.8 Flash} model, and 6-hour budget, but may train and ensemble any machine learning models without \tabfm{} (Table~\ref{tab:free_agent_ablation}). \texttt{Coding Agent} outperforms individual tuned GBDTs by training and blending LightGBM, CatBoost, random forests, linear models, and neural networks (Appendix~\ref{sec:app_free_agent}, Figure~\ref{fig:free_agent_models}). However, it is slightly worse than 4-hour \texttt{AutoGluon 1.3} and $510.8$ Elo below \tabfmauto{} (1468.8 vs.\ 1979.6): \tabfm{}'s pretrained prior adds $+316.5$ Elo (to 1785.3) and pipeline search adds $+194.3$ Elo. This ablation supports our core claim that coding agents perform better when paired with a frozen foundation model, since model architectures and training hyperparameters form a wide search space and trained models vary from run to run.

\paragraph{Transfer to Other Tabular Foundation Models.}
The agents choose every pipeline by how well \tabfm{} scores with it, which may make the final pipelines specific to \tabfm{}. We test this by running the four functions of the final pipelines $P^*$ from the three highest-rated configurations, unchanged and without further search, around three other frozen tabular foundation models (\texttt{TabPFN-3}, \texttt{TabICLv2}, and \texttt{EXAONE-Tabular}) in place of \tabfm{} (Appendix~\ref{sec:app_transfer}). From \texttt{TABFM\_KWARGS}, we keep only the context size, as the other models do not support settings such as NNLS view weighting, feature crosses, and SVD features. The pipelines of all three configurations improve every model over its default configuration, though by less than \tabfm{} does. Those from \texttt{Claude Code} with \texttt{Opus 5} transfer best (Figure~\ref{fig:transfer} and Appendix Figure~\ref{fig:transfer_arms}), raising \texttt{TabICLv2} by $+143.3$ Elo, \texttt{TabPFN-3} by $+130.8$ Elo, and \texttt{EXAONE-Tabular} by $+88.7$ Elo. The discovered pipelines therefore carry over to other in-context learners.

\input{tables/token_consumption}

\paragraph{Token and Tool Usage Across Harnesses and Models.}
Table~\ref{tab:token_consumption} summarizes evaluation counts, token usage, and tool-call distributions across the five configurations. We observe that \texttt{Opus 5} and \texttt{Gemini 3.8 Flash} reach similar test error with different search styles: \texttt{Opus 5} uses more tokens per step for reasoning and evaluates roughly one-third as many candidate pipelines under \texttt{Codex}, whereas \texttt{Gemini 3.8 Flash} makes $4.4\times$ to $5.2\times$ more tool calls to test rapid, incremental edits. The harnesses differ in \textit{Task \& Sched.} calls because their shell tools either block until an evaluation finishes (\texttt{Claude Code}) or poll background processes (\texttt{Codex} and \texttt{Antigravity}).

\subsection{MLE-Bench-Tabular}
\label{sec:exp_mlebench}

\begin{figure}[ht]
    \centering
    \includegraphics[width=\linewidth]{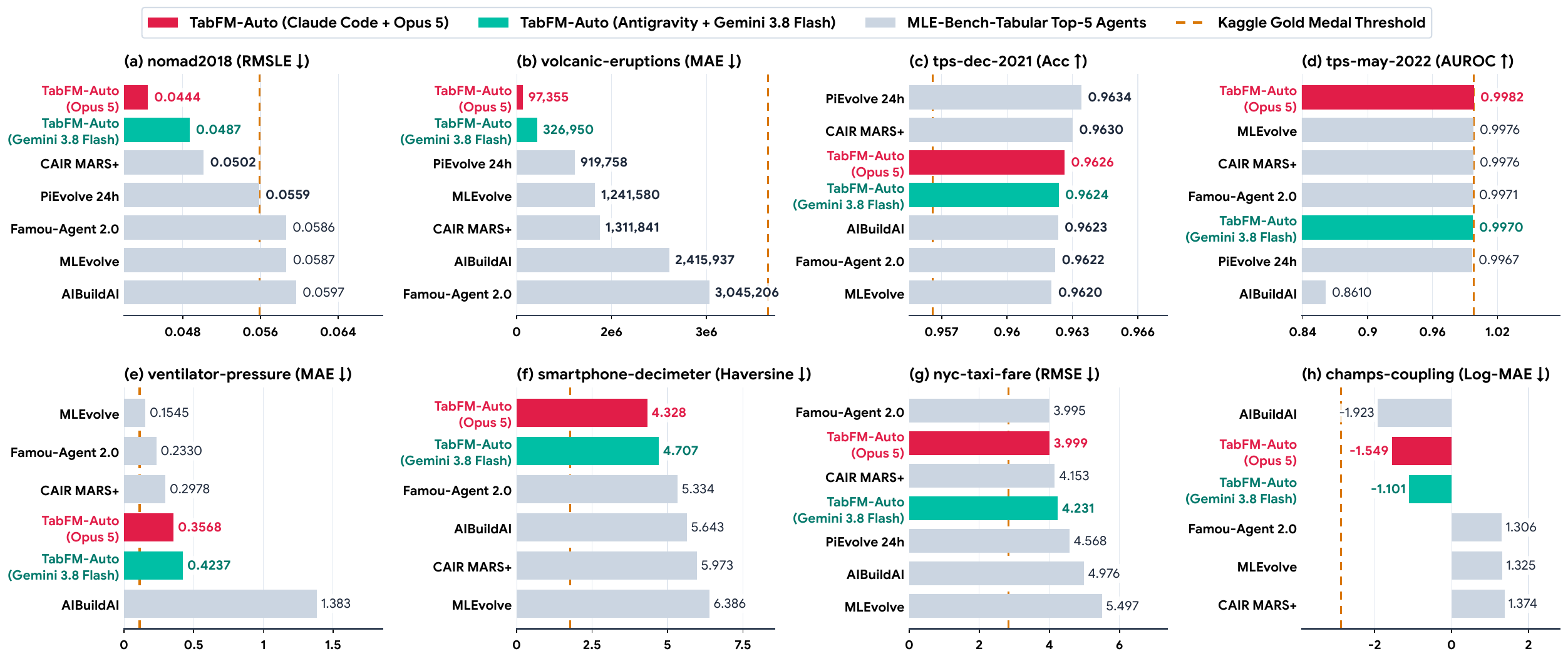}
    \caption{\textbf{Official test scores across all 8 MLE-Bench-Tabular competitions} comparing both \tabfmauto{} configurations against five leading external agents (CAIR MARS+, PiEvolve 24h, MLEvolve, Famou-Agent 2.0, and AIBuildAI) and Kaggle Gold Medal thresholds.}
    \label{fig:mlebench_leaderboard}
\end{figure}

\begin{figure}[t]
    \centering
    \includegraphics[width=\linewidth]{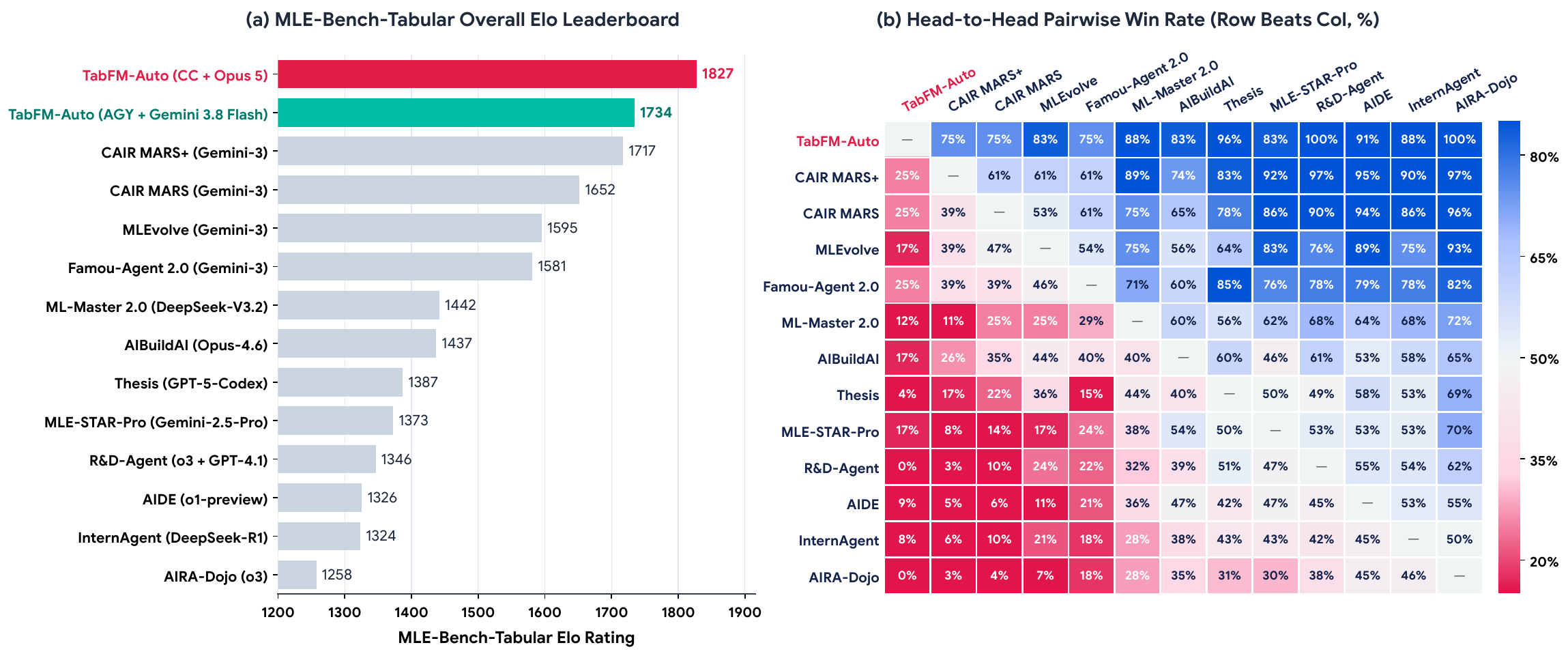}
    \caption{\textbf{MLE-Bench-Tabular overall comparison.} (a) Elo ratings of 14 MLE agents (our two configurations and all 12 external agents with complete 8-competition submissions, see Appendix~\ref{sec:app_mlebench_details}) and (b) pairwise win rates over the same run-vs-run matches showing the \texttt{Opus 5} variant.}
    \label{fig:mlebench_elo}
\end{figure}

\paragraph{Benchmark Definition and Auxiliary File Handling.}
We also evaluate \tabfmauto{} on MLE-Bench-Tabular, the 8 tabular competitions of MLE-Bench~\citep{chan2024mlebench} (Figure~\ref{fig:mlebench_leaderboard} and Appendix Table~\ref{tab:mlebench_results}), on the official local test splits (using a 24-hour per-competition budget for \tabfmauto{} against the official leaderboard submissions of external agents). In several competitions, the main signal is in auxiliary files such as seismic waveforms, satellite telemetry, or 3D molecular coordinates, which a single-table predictor cannot read directly. In $\Phi_{\text{feat}}$, \tabfmauto{} summarizes these files into tabular features (e.g., spectral energy or inter-atomic distances) and joins them to the main table before calling \tabfm{} (Appendix~\ref{sec:app_mlebench_details}).

\paragraph{Comparison with End-to-End Agents on MLE-Bench-Tabular.}
External MLE-Bench agents, such as CAIR MARS+, MLEvolve, Famou-Agent 2.0, R\&D-Agent, and AIDE, search over both data processing and task-specific predictive models, whereas \tabfmauto{} adapts the data pipeline while keeping the predictor fixed. Although external agents lead on several individual competitions (Figure~\ref{fig:mlebench_leaderboard}), \tabfmauto{} achieves the highest aggregate Elo among the evaluated submissions. Across all graded pairwise matches against the 12 external MLE agents (Figure~\ref{fig:mlebench_elo}), \tabfmauto{} ranks first and second overall (\texttt{Claude Code} with \texttt{Opus 5} at 1827 Elo, earning 4 Kaggle medals, and \texttt{Antigravity} with \texttt{Gemini 3.8 Flash} at 1734 Elo). \texttt{Opus 5} wins $86.7\%$ of all graded matches, including $75\%\text{--}88\%$ against each of the five highest-rated external agents and $83\%\text{--}100\%$ against the rest (Figure~\ref{fig:mlebench_elo}b).

%% file: tables/free_agent_ablation.tex
\begin{table}[t]
\centering
\caption{\textbf{Controlled ablation of \tabfmauto{} vs.\ an unconstrained coding agent.} Both agents use the identical harness, model, sandbox, and a 6-hour search budget per dataset.}
\label{tab:free_agent_ablation}
\resizebox{\linewidth}{!}{%
\setlength{\tabcolsep}{10pt}
\renewcommand{\arraystretch}{1.20}
\begin{tabular}{llccc}
\toprule
\textbf{Method} & \textbf{Working Models} & \textbf{Overall Elo} $\uparrow$ & \textbf{Cls.\ Elo} $\uparrow$ & \textbf{Reg.\ Elo} $\uparrow$ \\
\midrule
XGBoost (tuned + ensemble) & XGBoost & 1360.4 & 1361.7 & 1428.7 \\
LightGBM (tuned + ensemble) & LightGBM & 1420.1 & 1419.6 & 1511.7 \\
\rowcolor{gray!12}Coding Agent (AGY + Gemini 3.8 Flash) & Any ML models (GBDTs, PyTorch, etc.) & 1468.8 & 1472.0 & 1615.3 \\
AutoGluon 1.3 (4h) & Stacked GBDT \& NN ensembles & 1490.3 & 1473.4 & 1646.1 \\
TabFM & TabFM (frozen) & 1785.3 & 1768.7 & 2045.9 \\
TabFM+ & TabFM (frozen) & 1856.0 & 1836.7 & 2169.2 \\
\rowcolor{tabfmautored!12}TabFM-Auto (AGY + Gemini 3.8 Flash) & TabFM (frozen) & \textbf{1979.6} & \textbf{1937.5} & \textbf{2347.8} \\
\bottomrule
\end{tabular}%
}
\end{table}

%% file: tables/token_consumption.tex
\begin{table}[t]
\centering
\caption{\textbf{Cumulative evaluation count, API calls, token consumption, and tool invocation breakdown across 51 TabArena datasets.} (Top) Pipeline evaluations, LLM API requests, and prompt/generated token counts. (Bottom) Tool-call breakdown across the five action categories.}
\label{tab:token_consumption}
\label{tab:tool_calls_breakdown}
\setlength{\tabcolsep}{2.5pt}
\footnotesize
\begin{tabular*}{\linewidth}{@{\extracolsep{\fill}}lccccc@{}}
\toprule
\textbf{Agent Harness \& Model} & \textbf{Datasets} & \textbf{\# Evals} & \textbf{\# API Calls} & \textbf{Prompt Tokens} & \textbf{Generated Tokens} \\
\midrule
Codex, Opus 5 & 51 & 1,257 & 18,948 & 1.95B & 12.23M \\
Claude Code, Opus 5 & 51 & 4,296 & 7,452 & 1.28B & 9.76M \\
Antigravity, Gemini 3.8 Flash & 51 & 4,320 & 51,968 & 5.30B & 20.50M \\
Claude Code, Gemini 3.8 Flash & 51 & 4,602 & 38,401 & 3.75B & 9.07M \\
Codex, Gemini 3.8 Flash & 51 & 3,669 & 85,862 & 12.13B & 20.25M \\
\bottomrule
\end{tabular*}

\smallskip

\setlength{\tabcolsep}{1.5pt}
\scriptsize
\begin{tabular*}{\linewidth}{@{\extracolsep{\fill}}lcccccc@{}}
\toprule
\textbf{Agent Harness \& Model} & \textbf{Total Calls} & \textbf{Shell Exec.} & \textbf{File Read} & \textbf{File Write} & \textbf{File Search} & \textbf{Task \& Sched.} \\
\midrule
Codex, Opus 5 & 19,348 & 9,713 (50.2\%) & 270 (1.4\%) & 1,539 (8.0\%) & 120 (0.6\%) & 7,706 (39.8\%) \\
Claude Code, Opus 5 & 7,381 & 6,385 (86.5\%) & 256 (3.5\%) & 690 (9.3\%) & 32 (0.4\%) & 18 (0.2\%) \\
Antigravity, Gemini 3.8 Flash & 49,352 & 25,467 (51.6\%) & 5,473 (11.1\%) & 8,688 (17.6\%) & 1,455 (2.9\%) & 8,269 (16.8\%) \\
Claude Code, Gemini 3.8 Flash & 38,063 & 16,288 (42.8\%) & 15,398 (40.5\%) & 5,743 (15.1\%) & 611 (1.6\%) & 23 (0.1\%) \\
Codex, Gemini 3.8 Flash & 85,389 & 32,630 (38.2\%) & 3,066 (3.6\%) & 3,963 (4.6\%) & 3,669 (4.3\%) & 42,061 (49.3\%) \\
\bottomrule
\end{tabular*}
\end{table}

%% file: sections/05Conclusion.tex
\section{Conclusion and Discussion}
\label{sec:conclusion}

We studied how a language model coding agent can evolve the data pipeline around a frozen tabular foundation model, combining language-model domain reasoning with fast forward-pass tabular inference to supply the semantics missing from synthetic pretraining. On all 51 TabArena datasets, our five \tabfmauto{} configurations hold the top five overall positions (reaching 2013.0 Elo, $+227.7$ over \tabfm{}). The discovered pipelines transfer without further search to other frozen tabular foundation models ($+68.8$ to $+143.3$ Elo). \tabfmauto{} also ranks first overall on MLE-Bench-Tabular ahead of agents that train trees and neural networks from scratch.

Two limitations suggest directions for future work. Gains are generally smaller on anonymized tables, where the agent relies on statistical and relational features, and pipeline search requires repeated validation evaluations for each dataset. The discovered pipelines are short, readable programs. Operations collected across datasets could form a reusable library that warm-starts search on new tables with fewer evaluations. Incorporating the discovered formulas, cleaning rules, target transforms, and relational features into pretraining may lead to schema-aware tabular foundation models. Since \tabfmauto{} already joins features derived from auxiliary files on MLE-Bench-Tabular, the same search could extend to multi-table relational databases.

%% file: sections/06Appendix.tex
\section{Verification Protocol and Test Isolation}
\label{sec:app_verification}
\suppressfloats[t]

The evaluation harness enforces three checks during search and testing so that the error reductions reported for \tabfmauto{} come from the discovered data pipeline rather than from reading held-out test splits or from replacing \tabfm{} with another model.

\paragraph{Identity Baseline and Frozen-Model Check.}
Every search run begins by evaluating the identity pass-through pipeline $P_0 = (\mathrm{Id}, \mathrm{Id}, \mathrm{Id}, \mathrm{Id})$ (Figure~\ref{fig:identity_pipeline}) as its first evaluation (step~1). $P_0$ passes the unmodified table to \tabfm{} with default constructor settings on the same 3-fold cross-validation splits of fold~0's training set, and its score $\mathcal{U}_{\text{val}}(P_0)$ sets the starting threshold: the agent keeps only edits that improve the validation score over $P_0$. The harness runs the PyTorch checkpoint of \tabfm{} with a fixed ensemble of 8 members for every pipeline, so the test errors of $P_0$ differ slightly from the published \tabfm{} results in Table~\ref{tab:all_51_datasets}. The harness, not the pipeline, calls the frozen 400M-parameter \tabfm{} model $f_{\theta^*}$ on each candidate's transformed table and passes its predictions to \texttt{postprocess()}, so every final prediction goes through $f_{\theta^*}$.

\begin{figure}[ht]
\centering
\begin{minipage}{\linewidth}
\begin{lstlisting}[style=pybox,columns=fixed,mathescape=true]
import numpy as np

TABFM_KWARGS = {}  # TabFM constructor settings; {} keeps the defaults

def preprocess(X_train, y_train, X_test):        # $\Phi_{\text{clean}}$
  return X_train, y_train, X_test

def engineer(X_train, y_train, X_test):          # $\Phi_{\text{feat}}$
  return X_train, X_test

def sample(X_train, y_train, X_test, max_rows):  # $\mathcal{S}_{\text{ctx}}$
  return [np.arange(len(X_train))]  # one context view with all training rows

def postprocess(pred, y_train_orig, X_test):     # $\Psi_{\text{post}}$
  return pred
\end{lstlisting}
\end{minipage}
\caption{\textbf{Identity pipeline $P_0$ that starts every search run} (the starter \texttt{pipeline.py}, simplified). Every function returns its inputs unchanged and \texttt{TABFM\_KWARGS} is empty. The first evaluation thus scores \tabfm{} with default constructor settings on the raw table. The harness fits the frozen \tabfm{} (8 ensemble members, configured by \texttt{TABFM\_KWARGS}) once per context view returned by \mbox{\texttt{sample()}}, averages the test predictions over views, and passes them to \texttt{postprocess()}, where \mbox{\texttt{y\_train\_orig}} is the untransformed training target used to invert target transforms. The hooks also receive a \texttt{row\_id} column, which the harness drops before calling \tabfm{}. \texttt{sample()} is optional, and the listing writes out its default. \texttt{engineer()} may return at most 500 columns, so $P_0$ fails on the three datasets with more columns (\texttt{Bioresponse}, \texttt{hiva\_agnostic}, and \texttt{QSAR-TID-11}).}
\label{fig:identity_pipeline}
\end{figure}

\paragraph{Operating-System Sandboxing and Split Isolation.}
Each TabArena dataset defines 9 or 30 official train/test evaluation splits (3-fold cross-validation repeated 3 or 10 times). During pipeline search, the evaluation harness gives the agent access only to fold~0's training split, partitioning it into 3 internal cross-validation folds $(\mathcal{D}_{\text{subtrain}}, \mathcal{D}_{\text{val}})$ to compute $\mathcal{U}_{\text{val}}$. The harness runs every candidate pipeline in an unprivileged bubblewrap Linux namespace sandbox that disables outbound network access, mounts system libraries read-only, and restricts writes to an ephemeral scratch directory. The sandbox also unmounts fold~0's official test split along with the train and test splits of all other folds across the 51 datasets. Only after search finishes and $P^*$ is frozen does the harness score $P^*$ once on those held-out splits in an isolated environment.

\paragraph{Row-Order Permutation Against Index Leakage.}
The raw files of several public classification datasets are sorted by target class or acquisition timestamp, so a pipeline that preserves the original row order could infer labels from row positions. The harness therefore deterministically shuffles all classification folds at load time and restores the original order only when scoring final predictions.

\section{Full TabArena Results and Taxonomy Analysis}
\label{sec:app_tabarena}

\input{tables/leaderboard_overall}
\input{tables/leaderboard_fold0}

\begin{figure}[t]
    \centering
    \includegraphics[width=\linewidth]{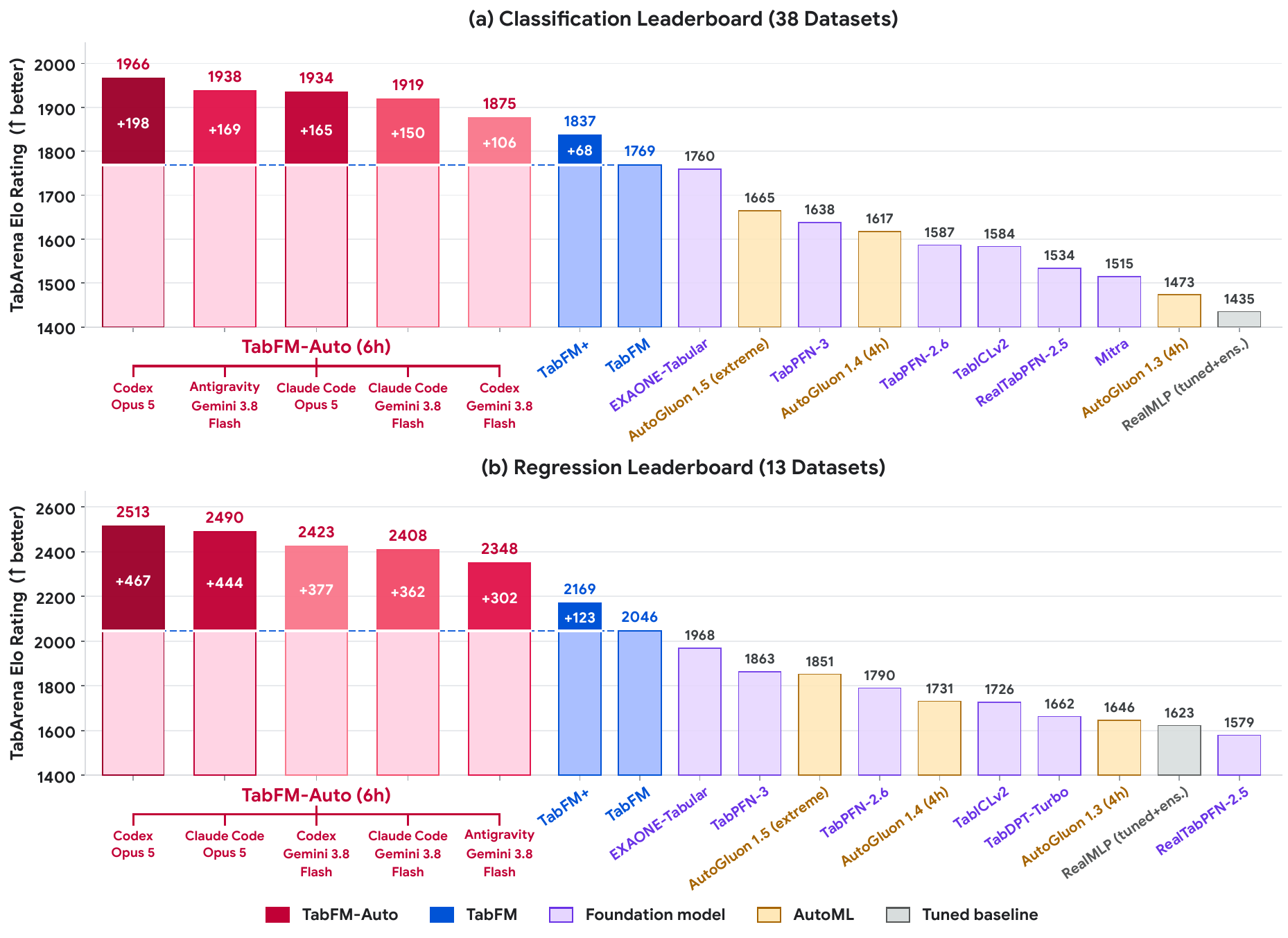}
    \caption{\textbf{TabArena Elo ratings, computed separately on the 38 classification (top) and 13 regression (bottom) datasets.} Red bars denote \tabfmauto{} configurations (\#1--\#5), while blue bars denote \tabfmp{} (\#6) and \tabfm{} (\#7).}
    \label{fig:leaderboards}
\end{figure}

\input{tables/feature_engineering_taxonomy}

\subsection{Overall Leaderboard and Baseline Comparison}
\label{sec:app_leaderboard_analysis}

Table~\ref{tab:overall} and Figure~\ref{fig:leaderboards} report the full leaderboard across all 51 TabArena datasets, comparing our five \tabfmauto{} configurations against tuned GBDTs, deep tabular models, 4-hour \texttt{AutoGluon 1.5}, and other tabular foundation models. \tabfm{} and \tabfmp{} already outperform individual GBDT families and 4-hour \texttt{AutoGluon 1.5} (1668.4 Elo). Evolving the data pipeline around frozen \tabfm{} reduces overall oracle improvability from $3.20\%$ (\tabfm{}) and $2.20\%$ (\tabfmp{}) down to $1.26\%\text{--}1.88\%$ and more than doubles the number of dataset wins (from $9.74$ for \tabfm{} to $26.12$ for \texttt{Antigravity} with \texttt{Gemini 3.8 Flash} and $25.00$ for \texttt{Codex} with \texttt{Opus 5}).

\begin{figure}[t]
    \centering
    \includegraphics[width=\linewidth]{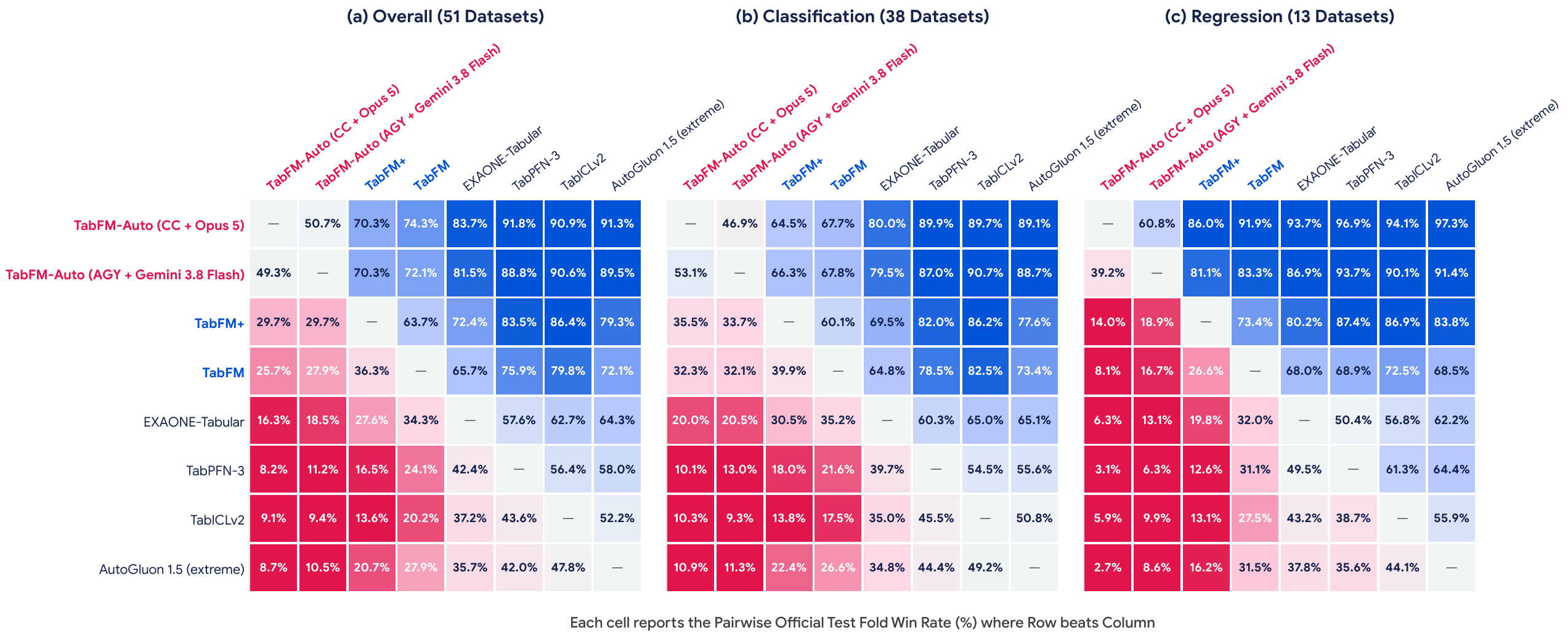}
    \caption{\textbf{Pairwise official test fold win rates on TabArena} across (a) all 51 datasets, (b) 38 classification datasets, and (c) 13 regression datasets. Each cell reports the percentage of dataset-fold instances where the row method achieves lower official test error than the column method.}
    \label{fig:tabarena_win_rate}
\end{figure}

\paragraph{Fold-0 Held-Out Evaluation and Cross-Fold Pipeline Transfer.}
\tabfmauto{} searches for a single Python data pipeline script $P^* = (\Phi_{\text{clean}}, \Phi_{\text{feat}}, \mathcal{S}_{\text{ctx}}, \Psi_{\text{post}})$ per dataset using 3-fold cross-validation on Fold~0's training split $\mathcal{D}_{\text{train}}^{(0)}$ and evaluates $P^*$ across all $S \in \{9, 30\}$ official evaluation folds $k \in \{0, \dotsc, S-1\}$. Because TabArena repeats 3-fold cross-validation, the test sets $\mathbf{X}_{\text{test}}^{(k)}$ of Folds $k \ge 1$ share rows with $\mathcal{D}_{\text{train}}^{(0)}$, so selecting $P^*$ on $\mathcal{D}_{\text{train}}^{(0)}$ could inflate their test scores. Three aspects of our protocol and results are relevant here:
\begin{enumerate}[leftmargin=16pt, itemsep=2pt, topsep=2pt]
    \item \textit{One search per dataset:} As in manual feature engineering, the agent designs a single dataset-level pipeline (which domain ratios to compute, which sentinel values to mask, and whether to transform the target) and reuses it across every train/test split, rather than writing 9 or 30 different feature sets for the same dataset. Running an independent 6-hour agent search on all 816 folds across five configurations ($4{,}080$ runs) would increase LLM token and GPU compute costs by $16\times$ (from $1.28\text{B}\text{--}12.13\text{B}$ prompt tokens per 51-dataset sweep in Table~\ref{tab:token_consumption} to $20.5\text{B}\text{--}194.1\text{B}$ tokens). The five sweeps in Table~\ref{tab:token_consumption} cost about \$17.6K in LLM API fees at Vertex AI list prices. This protocol lets the search see rows that later serve as test rows in Folds $k \ge 1$, so Fold~0 is the only fully held-out split.
    \item \textit{Symbolic program structure vs.\ fitted numerical parameters:} What transfers across folds is only a short Python source script (\texttt{pipeline.py}) specifying column-level operations, such as physical scaling laws ($St = f\delta^*/U_\infty$), sentinel rules ($\texttt{width}=0 \to \mathrm{NaN}$), and target link functions ($\log(1+y)$). On every evaluation fold $k \in \{0, \dotsc, S-1\}$, the harness executes \texttt{pipeline.py} in a fresh isolated process where all data-dependent statistics (imputers, frequency maps, SVD bases, empirical class priors) and \tabfm{}'s in-context conditioning are fitted only on fold~$k$'s training split $(\mathbf{X}_{\text{train}}^{(k)}, \mathbf{y}_{\text{train}}^{(k)})$ before predicting on fold~$k$'s held-out test split $\mathbf{X}_{\text{test}}^{(k)}$. No model weights, fitted statistics, or row predictions from $\mathcal{D}_{\text{train}}^{(0)}$ carry over: $\mathcal{D}_{\text{train}}^{(0)}$ affects Folds $k \ge 1$ only through which operations and settings the agent kept.
    \item \textit{Fold-0 held-out evaluation (Table~\ref{tab:fold0_elo}):} Scored on Fold~0's test split $\mathcal{D}_{\text{test}}^{(0)}$ alone, which shares no rows with $\mathcal{D}_{\text{train}}^{(0)}$, \tabfmauto{} holds the \#1 Overall and \#1 Classification ratings ($1950.7$ and $1877.5$ Elo, both \texttt{Codex} with \texttt{Opus 5}) and the \#1--\#4 Regression ratings ($2532.4\text{--}2645.6$ Elo, led by \texttt{Codex} with \texttt{Gemini 3.8 Flash}), and all five configurations outrank \texttt{EXAONE-Tabular} ($1772.2$) and 4-hour \texttt{AutoGluon 1.5} ($1661.0$) overall. For the five configurations, pairwise win rates against the 62 TabArena methods with complete results are $91.1\%\text{--}95.7\%$ on Fold~0 versus $94.4\%\text{--}95.9\%$ on Folds $k \ge 1$ ($95.7\%$ vs.\ $95.9\%$ for \texttt{Codex} with \texttt{Opus 5}, $93.6\%$ vs.\ $94.7\%$ for \texttt{Claude Code} with \texttt{Gemini 3.8 Flash}, and $93.0\%$ vs.\ $95.8\%$ for \texttt{Claude Code} with \texttt{Opus 5}). For \texttt{Codex} with \texttt{Opus 5}, the top-rated configuration in Table~\ref{tab:overall}, the G-Mean test error reduction over \tabfm{} on Fold~0 is close to that on Folds $k \ge 1$ ($3.74\%$ vs.\ $4.53\%$ on classification and $3.23\%$ vs.\ $3.28\%$ on regression).
\end{enumerate}

\subsection{Analysis of Discovered Pipelines}
\label{sec:app_taxonomy_analysis}
\label{sec:app_pipeline_stages}

We analyze the final pipelines $P^* = (\Phi_{\text{clean}}, \Phi_{\text{feat}}, \mathcal{S}_{\text{ctx}}, \Psi_{\text{post}})$ discovered by \texttt{Claude Code} with \texttt{Opus 5} and \texttt{Antigravity} with \texttt{Gemini 3.8 Flash} across all 51 TabArena datasets (Table~\ref{tab:feature_engineering_taxonomy}). We count how often the agents edit each stage and inspect the functions they write to see how they address the limitations of synthetic pretraining.

\paragraph{Data Cleaning and Target Conditioning ($\Phi_{\text{clean}}$).}
Both \texttt{Opus 5} and \texttt{Gemini 3.8 Flash} modify the cleaning and target-conditioning stage on $33.3\%$ of the benchmark ($17/51$ datasets each). Because \tabfm{} embeds continuous columns through shared linear projections before computing row-wise attention, uncleaned numerical placeholders (such as \texttt{0} for unmeasured steel coil width on \texttt{anneal} or \texttt{-1} and \texttt{999} on administrative records) distort row-wise attention weights. By checking column names and value histograms, the agents replace these placeholders with missing-value indicators. On regression datasets with skewed targets, $\Phi_{\text{clean}}$ also applies reversible transforms $g(\mathbf{y})$ such as $\log(1+y)$ or Box--Cox scaling before calling \tabfm{}, and $\Psi_{\text{post}}$ inverts them with $g^{-1}$ at the output.

\paragraph{I: Domain-Knowledge Features ($\Phi_{\text{feat}}$ on Informative Schemas, 17 Datasets).}
\texttt{Opus 5} modifies the feature table via $\Phi_{\text{feat}}$ (or $\Phi_{\text{clean}}$) on $100\%$ of datasets ($51/51$) and \texttt{Gemini 3.8 Flash} on $90.2\%$ ($46/51$), for a pooled rate of $95.1\%$ ($97/102$). On the 17 datasets whose column names or task descriptions identify real-world quantities, such as physical, clinical, engineering, or economic variables (Category~I in Table~\ref{tab:feature_engineering_taxonomy}), both models translate domain relationships directly into nonlinear composite features, reducing multi-fold official test error by $+7.25\%$ (\texttt{Opus 5}) and $+8.16\%$ (\texttt{Gemini 3.8 Flash}), and by $+8.85\%$ under validation-selected \textit{Combined Best}. These 17 datasets account for 6 of the 10 largest dataset improvements on TabArena for each configuration. These domain features fall into three groups where explicit metadata-guided features complement synthetic pretraining:
\begin{itemize}[leftmargin=16pt, itemsep=2pt, topsep=2pt]
    \item \textit{Physical Ratios and Engineering Formulas:} Alternating row-and-column attention approximates smooth additive combinations well, but benefits from explicit multiplicative ratios and trigonometric projections on physical tasks. On engineering datasets, the agents build dimensionless scaling terms directly from variable names (Figure~\ref{fig:fe_case_studies}-I). Examples include Strouhal numbers $St = f\delta^*/U_\infty$ and Helmholtz compactness $He = fc/a_0$ on \texttt{airfoil\_self\_noise} ($+14.6\%$ and $+17.3\%$ test RMSE reduction for \texttt{Opus 5} and \texttt{Gemini 3.8 Flash}, respectively), water-to-binder ratios $W / (C + 0.8S + 0.3F)$ on \texttt{concrete\_strength} ($+3.8\%$, \texttt{Opus 5}), and mass-weighted thermal conductivity ratios on \texttt{superconductivity}.
    \item \textit{Clinical Diagnostic Codes and Vital-Sign Ratios:} On medical records where categorical columns store alphanumeric diagnostic codes, a tabular foundation model treats each code as an unrelated discrete token. On \texttt{Diabetes130US}, the agents parse 3-digit ICD-9 strings into 19 physiological organ-system chapters with dedicated offsets for supplementary V-codes and E-codes, coarsening hundreds of sparse categories into clinically meaningful strata. On intensive-care and obstetric cohorts (\texttt{MIC} and \texttt{maternal\_health\_risk}), the agents combine systolic blood pressure (SBP), diastolic blood pressure (DBP), and heart rate (HR) into standard clinical indices (Mean Arterial Pressure $\mathrm{MAP} = (2\mathrm{DBP} + \mathrm{SBP})/3$, Pulse Pressure $\mathrm{SBP} - \mathrm{DBP}$, and Shock Index $\mathrm{HR}/\mathrm{SBP}$).
    \item \textit{Domain Sequence Motifs and Spectral Color Indices:} On scientific tables encoding sequences or multi-band fluxes, the agents construct domain-specific local features: extracting position-specific dinucleotide and trinucleotide sequence motifs around the donor/acceptor junction on \texttt{splice} ($+22.2\%$ test log-loss reduction, \texttt{Opus 5}) and forming pairwise photometric color indices ($u-g, g-r, r-i, i-z$) across adjacent filter bands on Sloan Digital Sky Survey observations (\texttt{SDSS17}, $+18.8\%$ log-loss reduction, \texttt{Gemini 3.8 Flash}).
\end{itemize}

\paragraph{II: Statistical and Structural Features ($\Phi_{\text{feat}}$ on Generic Schemas, 34 Datasets).}
On the remaining 34 datasets with anonymized headers (\texttt{f\_01}--\texttt{f\_N}) or generic business counters (Category~II in Table~\ref{tab:feature_engineering_taxonomy}), the agents cannot infer domain formulas from column names. Instead, the agents use $\Phi_{\text{feat}}$ to handle inputs that tabular transformers model poorly, achieving multi-fold test error reductions of $+2.31\%$ (\texttt{Opus 5}, $34/51$ datasets) and $+2.99\%$ (\texttt{Gemini 3.8 Flash}, $29/51$ datasets), and $+3.87\%$ under \textit{Combined Best}. Three kinds of features dominate Category~II:
\begin{itemize}[leftmargin=16pt, itemsep=2pt, topsep=2pt]
    \item \textit{Bipartite Graph Degree and Co-occurrence Features:} When a table consists of high-cardinality categorical entity IDs (such as employee roles, departments, managers, and resource IDs on \texttt{Amazon\_employee\_access}), standard categorical encodings assign arbitrary integer indices that hide which entities co-occur. Both agents compute label-free features on the combined train-and-test entity graph (entity frequencies, pairwise co-occurrences, and bipartite node degrees), reducing official test error $1-\mathrm{AUROC}$ (where $\mathrm{AUROC}$ is the area under the receiver operating characteristic curve, and lower error is better) from $0.1416$ to $0.1039$ (\texttt{Opus 5}) and $0.1063$ (\texttt{Gemini 3.8 Flash}) ($+26.6\%$ and $+25.0\%$ relative error reduction).
    \item \textit{Frequency Encodings and Missingness Signals:} On \texttt{kddcup09\_appetency}, whose 212 anonymized columns include 38 high-cardinality string columns and many columns that are mostly missing, \texttt{Opus 5} encodes each string column by its level count and a cross-fitted target mean, ranks the engineered columns by in-fold univariate $\mathrm{AUROC}$ with missing values treated as their own extreme (so that missingness counts as signal), and keeps only the top-ranked columns.
    \item \textit{Pruning Redundant Columns and SVD on Wide Tables:} On wide cheminformatics and bioassay matrices (\texttt{Bioresponse} and \texttt{hiva\_agnostic}) where over a thousand sparse molecular or bioassay columns dilute column-wise attention, the agents prune duplicate and near-collinear columns and append low-rank truncated SVD projections that summarize the highest-variance directions in a few columns.
\end{itemize}

\paragraph{Context Selection ($\mathcal{S}_{\text{ctx}}$).}
\texttt{Opus 5} and \texttt{Gemini 3.8 Flash} modify the context-selection stage $\mathcal{S}_{\text{ctx}}$ on $25.5\%$ ($13/51$) and $23.5\%$ ($12/51$) of datasets, respectively. When training tables exceed \tabfm{}'s 16,384-row pretraining length or exhibit severe class imbalance, uniform subsampling drops rare positive instances. The agents avoid this by returning several context index sets $\{I_1, \dotsc, I_K\}$: one drawn from the natural class distribution and others that oversample the minority class or stratify by cluster.

\paragraph{III: Output Calibration ($\Psi_{\text{post}}$) and Constructor Settings (\texttt{TABFM\_KWARGS}).}
Finally, \texttt{Opus 5} and \texttt{Gemini 3.8 Flash} tune input normalization and view weighting in \texttt{TABFM\_KWARGS} on $86.3\%$ and $98.0\%$ of datasets, respectively, and adjust post-hoc output calibration in $\Psi_{\text{post}}$ on $41.2\%$ ($21/51$) and $49.0\%$ ($25/51$). One possible reason is that the synthetic prior used to pretrain \tabfm{} need not match the class balance and confidence levels of real datasets. In $\Psi_{\text{post}}$, the agents shift predicted probabilities toward the empirical training class prior $\hat{\boldsymbol{\pi}}_{\text{train}}$ in log-odds space and apply temperature scaling. Category~III in Table~\ref{tab:feature_engineering_taxonomy} contains the 5 runs (all in \texttt{Gemini 3.8 Flash}, as \texttt{Opus 5} modifies the feature table on all 51 datasets) where the agent does not change the feature matrix ($\Phi_{\text{clean}} = \Phi_{\text{feat}} = \mathrm{Id}$), averaging $+2.16\%$ test error reduction across the 5 datasets: 4 of the 5 adjust context sampling in $\mathcal{S}_{\text{ctx}}$ and/or log-odds prior calibration in $\Psi_{\text{post}}$ (such as \texttt{E-CommereShippingData} at $+6.94\%$ and \texttt{website\_phishing} at $+3.82\%$), whereas the single run that tuned only constructor hyperparameters scored $-0.30\%$ on test.

\subsection{Transferring Final Pipelines to Other Tabular Foundation Models}
\label{sec:app_transfer}

\begin{figure}[t]
    \centering
    \includegraphics[width=\linewidth]{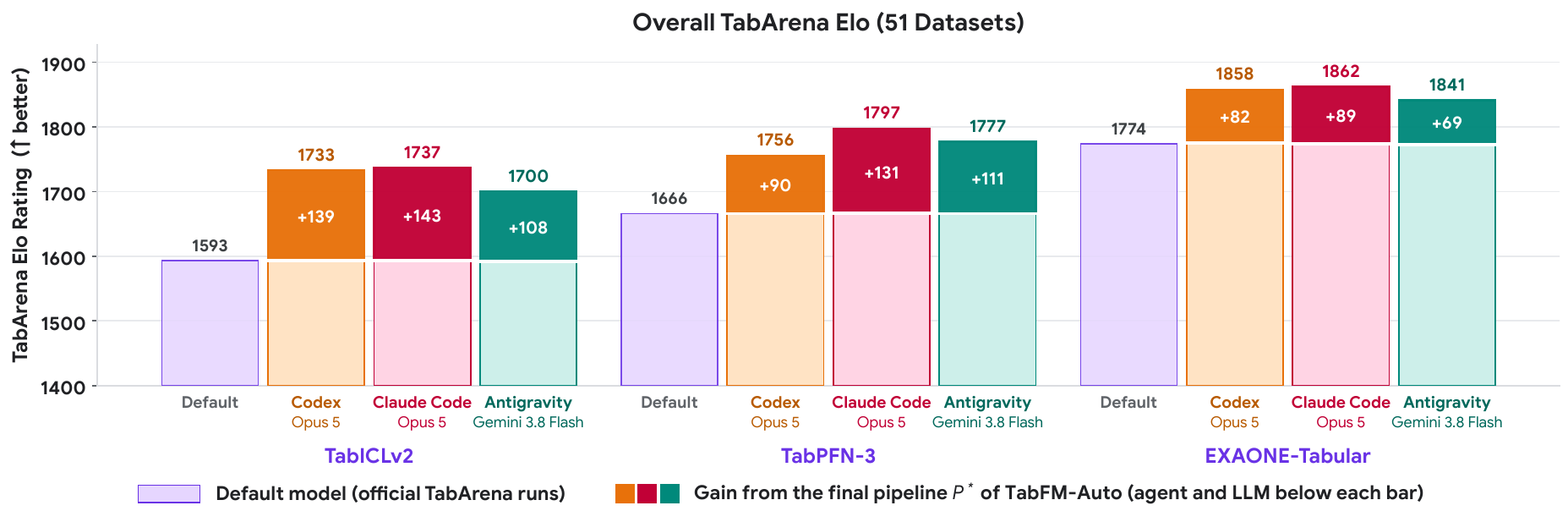}
    \caption{\textbf{The final pipelines of all three configurations help other tabular foundation models, and those from \texttt{Claude Code} with \texttt{Opus 5} help most.} Overall TabArena Elo on the 51 datasets of each model with its default configuration and within the final pipelines $P^*$ that each \tabfmauto{} configuration found for \tabfm{}. Each model with $P^*$ is rated in its own fit: the pale part of the bar reaches the default model rated in that fit, and the colored part is the gain. \emph{Default} is the mean of the three default ratings.}
    \label{fig:transfer_arms}
\end{figure}

\begin{figure}[t]
    \centering
    \includegraphics[width=\linewidth]{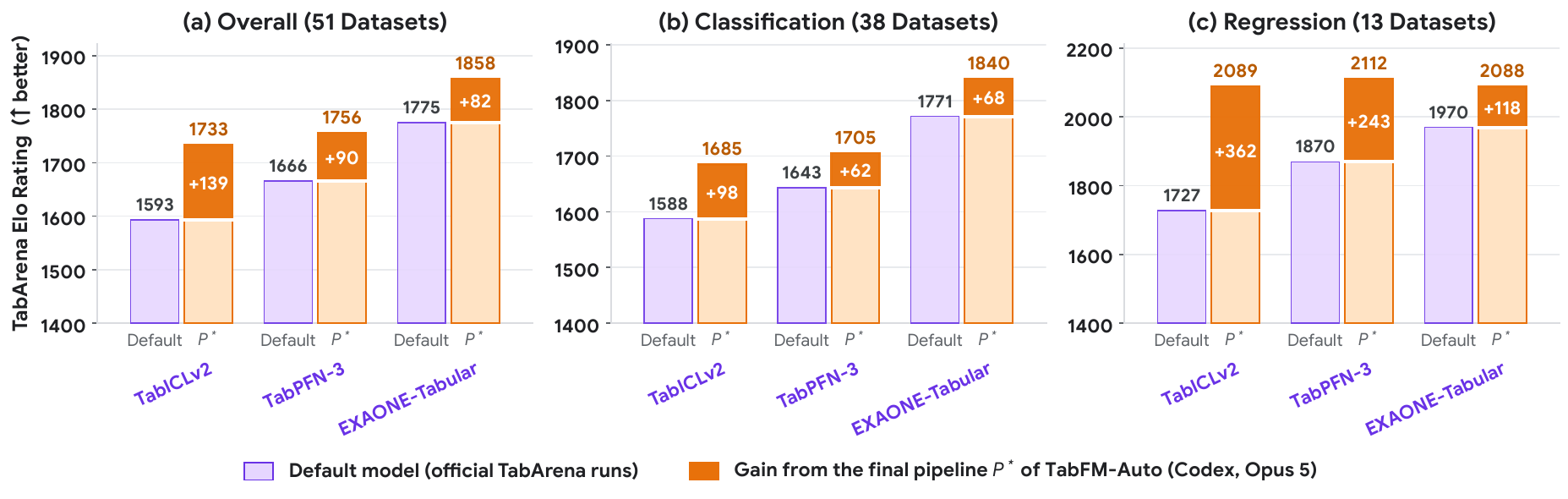}\\[4pt]
    \includegraphics[width=\linewidth]{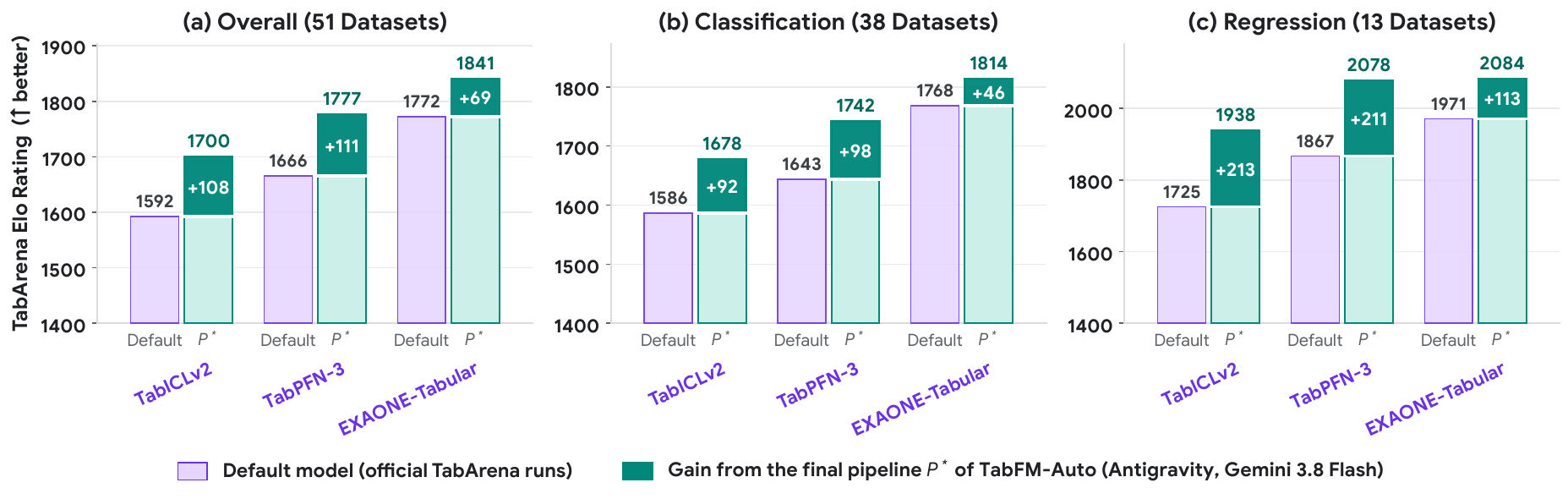}
    \caption{\textbf{The final pipelines from the other two configurations also help every model.} As in Figure~\ref{fig:transfer}, with the pipelines that \tabfmauto{} found with \texttt{Codex} and \texttt{Opus 5} (top) and with \texttt{Antigravity} and \texttt{Gemini 3.8 Flash} (bottom). \emph{Default} is the default model rated in the same fit as $P^*$.}
    \label{fig:transfer_other}
\end{figure}

For each dataset, we take the final pipeline $P^*$ that \texttt{Codex} with \texttt{Opus 5}, \texttt{Claude Code} with \texttt{Opus 5}, or \texttt{Antigravity} with \texttt{Gemini 3.8 Flash} found for \tabfm{}, and run its four functions unchanged around another frozen tabular foundation model. Each ensemble member keeps the number of context rows that \tabfm{} used with that pipeline. We do not carry over the other \texttt{TABFM\_KWARGS} settings, because the other models do not support many of them, such as NNLS view weighting, feature crosses, and SVD features. We score every official test fold of all 51 datasets and compare with the official TabArena results of each model's default configuration. Each model with $P^*$ from one configuration enters the Elo pool of Table~\ref{tab:overall} on its own and is rated in the same fit as its default version, so default ratings differ slightly from Table~\ref{tab:overall} and between configurations. Table~\ref{tab:transfer} lists the ratings, Figure~\ref{fig:transfer_arms} compares the three configurations on overall Elo, and Figure~\ref{fig:transfer_other} shows the pipelines of \texttt{Codex} with \texttt{Opus 5} and of \texttt{Antigravity} with \texttt{Gemini 3.8 Flash} by task type.

Figure~\ref{fig:transfer} shows the pipelines from \texttt{Claude Code} with \texttt{Opus 5}, which give every model its largest overall gain (+88.7 to +143.3 Elo). Those from \texttt{Codex} with \texttt{Opus 5} come close on \texttt{TabICLv2} (+139.4 vs.\ +143.3) and give the largest gains for \texttt{TabICLv2} on regression and \texttt{EXAONE-Tabular} on classification. Those from \texttt{Antigravity} with \texttt{Gemini 3.8 Flash} give the smallest overall gains on \texttt{TabICLv2} and \texttt{EXAONE-Tabular}. With the pipelines of any of the three configurations, every model improves on both task types, though by less than \tabfm{} gains from the same pipelines.

\input{tables/transfer_elo}

\subsection{Models Used by the Unconstrained Coding Agent}
\label{sec:app_free_agent}

\begin{figure}[t]
    \centering
    \includegraphics[width=\linewidth]{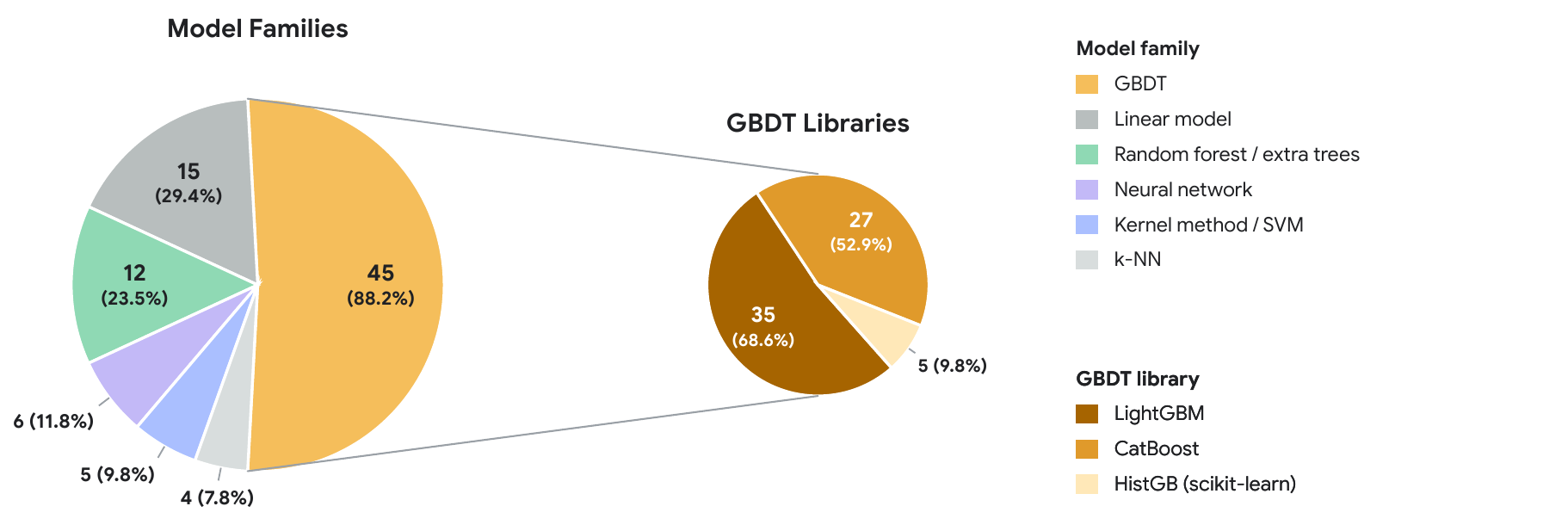}
    \caption{\textbf{Model families in the final pipelines of the unconstrained coding agent} (\texttt{Antigravity} with \texttt{Gemini 3.8 Flash}, no \tabfm{}). Each wedge gives the number of final pipelines, out of 51, whose code uses that family, and in parentheses that number as a percentage of the 51 pipelines. Since a pipeline that blends several families counts once for each, the percentages sum to more than 100\%. The right pie splits the GBDT wedge by library.}
    \label{fig:free_agent_models}
\end{figure}

Figure~\ref{fig:free_agent_models} shows how often each model family appears in the code of the 51 final pipelines written by the unconstrained coding agent of Table~\ref{tab:free_agent_ablation}. Gradient-boosted trees appear in 45 of them, mostly LightGBM (35) and CatBoost (27). No final pipeline uses XGBoost, and only 6 use a neural network. Twenty-three pipelines rely on a single model family. The other 28 blend two or more.

\begin{figure}[t]
    \centering
    \includegraphics[width=\linewidth]{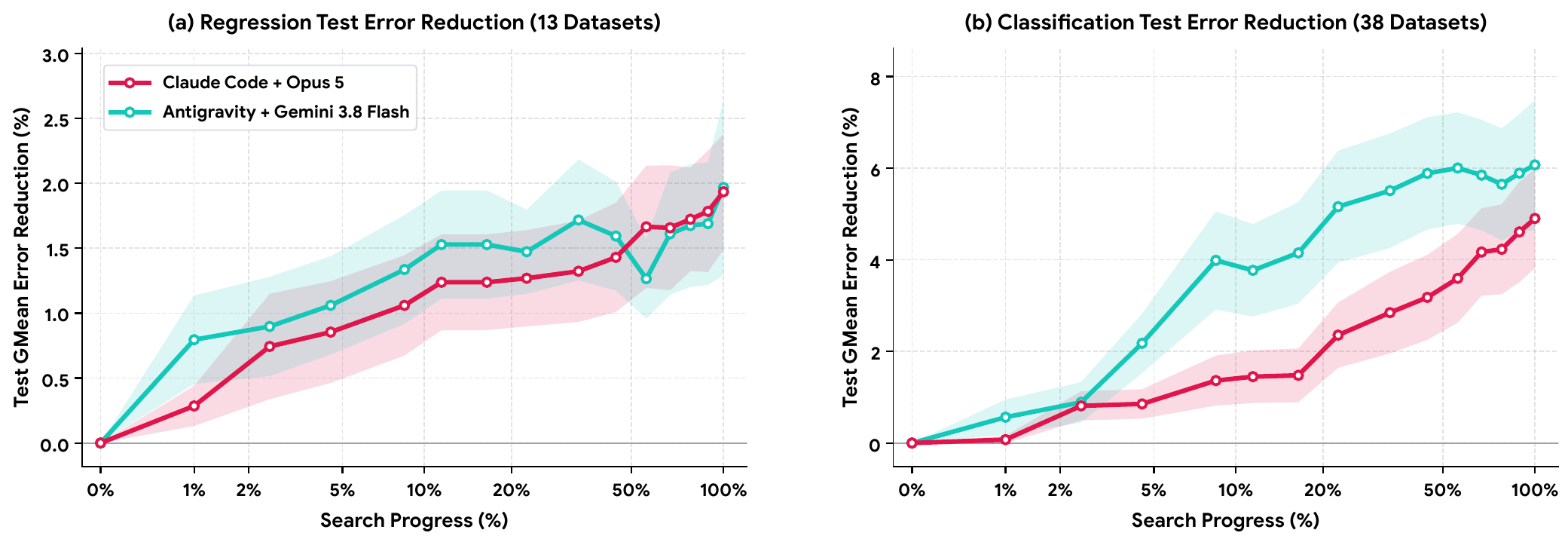}
    \caption{\textbf{Official test G-Mean error reduction across search progress (log scale).} (a) Regression suite (13 datasets) and (b) Classification suite (38 datasets).}
    \label{fig:hill_climbing_split}
\end{figure}

\section{Search Dynamics and Generalization Analysis}
\label{sec:app_dynamics}
\label{sec:app_generalization}

We saved every candidate pipeline evaluated during the TabArena runs of \texttt{Claude Code} with \texttt{Opus 5} and \texttt{Antigravity} with \texttt{Gemini 3.8 Flash}, and rescored up to 15 checkpoints per run, taken at fixed fractions of the search budget ($1{,}486$ checkpoints in total), on the official test sets across all folds to test whether repeated 3-fold cross-validation evaluations on fold~0's training set cause validation overfitting.

Figure~\ref{fig:hill_climbing_split} plots the official test G-Mean error reduction relative to $P_0$ separately for the 13 regression datasets (Figure~\ref{fig:hill_climbing_split}a) and the 38 classification datasets (Figure~\ref{fig:hill_climbing_split}b) over log-scaled search progress. On both suites, official test error across all folds decreases together with the search validation error from the first evaluation to the end of the budget ($100\%$ progress). On regression, both configurations make rapid early gains in the first $10\%$ of search as they add target transforms and primary physical ratios, then refine to $1.93\%$ (\texttt{Opus 5}) and $1.97\%$ (\texttt{Gemini 3.8 Flash}) lower G-Mean test RMSE than $P_0$ at full budget ($3.11\%$ and $3.15\%$ lower than the published \tabfm{} results in Table~\ref{tab:all_51_datasets}). On classification, official test error reduction climbs throughout the budget as agents add group aggregations, string decompositions, and prior calibration, reaching $4.90\%$ (\texttt{Opus 5}) and $6.07\%$ (\texttt{Gemini 3.8 Flash}) lower G-Mean test classification error than $P_0$ at $100\%$ of the budget.

\begin{figure}[t]
    \centering
    \includegraphics[width=\linewidth]{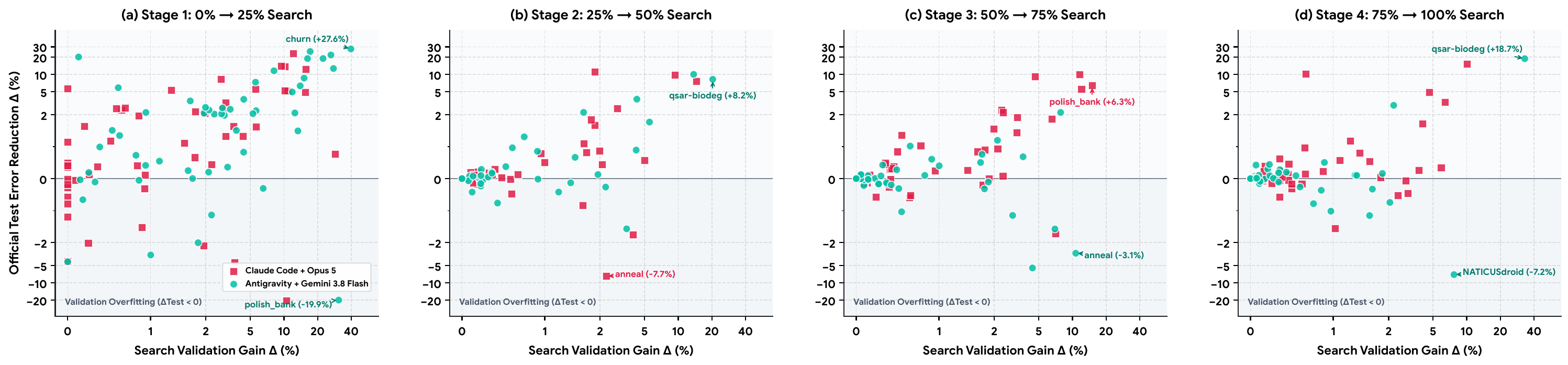}
    \caption{\textbf{Stage-wise validation-to-test generalization across the four search quarters (0\%--25\%, 25\%--50\%, 50\%--75\%, and 75\%--100\%).} Official test error reduction ($\Delta\%$) across all folds versus 3-fold cross-validation gain ($\Delta\%$) on fold~0's training set per dataset.}
    \label{fig:overfitting_symlog}
\end{figure}

In Figure~\ref{fig:overfitting_symlog}, we examine validation-to-test transfer within each quarter of the search budget ($0\%\text{--}25\%$, $25\%\text{--}50\%$, $50\%\text{--}75\%$, and $75\%\text{--}100\%$) by plotting marginal official test error reduction ($\Delta\%$) across all folds against marginal 3-fold cross-validation gain ($\Delta\%$) on fold~0's training set during that quarter for each dataset. In most quarters, most datasets with a validation gain also improve on test (upper-right quadrant), and datasets that have plateaued lie near the origin. The exception is the $50\%\text{--}75\%$ quarter of \texttt{Gemini 3.8 Flash}, where test changes split roughly evenly. In the final quarter ($75\%\text{--}100\%$ of budget), validation and test gains remain positively correlated across the datasets whose validation score still improves. Keeping \tabfm{} fixed and evaluating candidate edits via 3-fold cross-validation on $\mathcal{D}_{\text{train}}^{(0)}$ helps the discovered pipelines generalize across the held-out test folds.

\section{MLE-Bench-Tabular Analysis and Per-Competition Results}
\label{sec:app_mlebench}

\subsection{Competition-by-Competition Analysis on MLE-Bench-Tabular}
\label{sec:app_mlebench_details}

\begin{figure}[t]
    \centering
    \includegraphics[width=\linewidth]{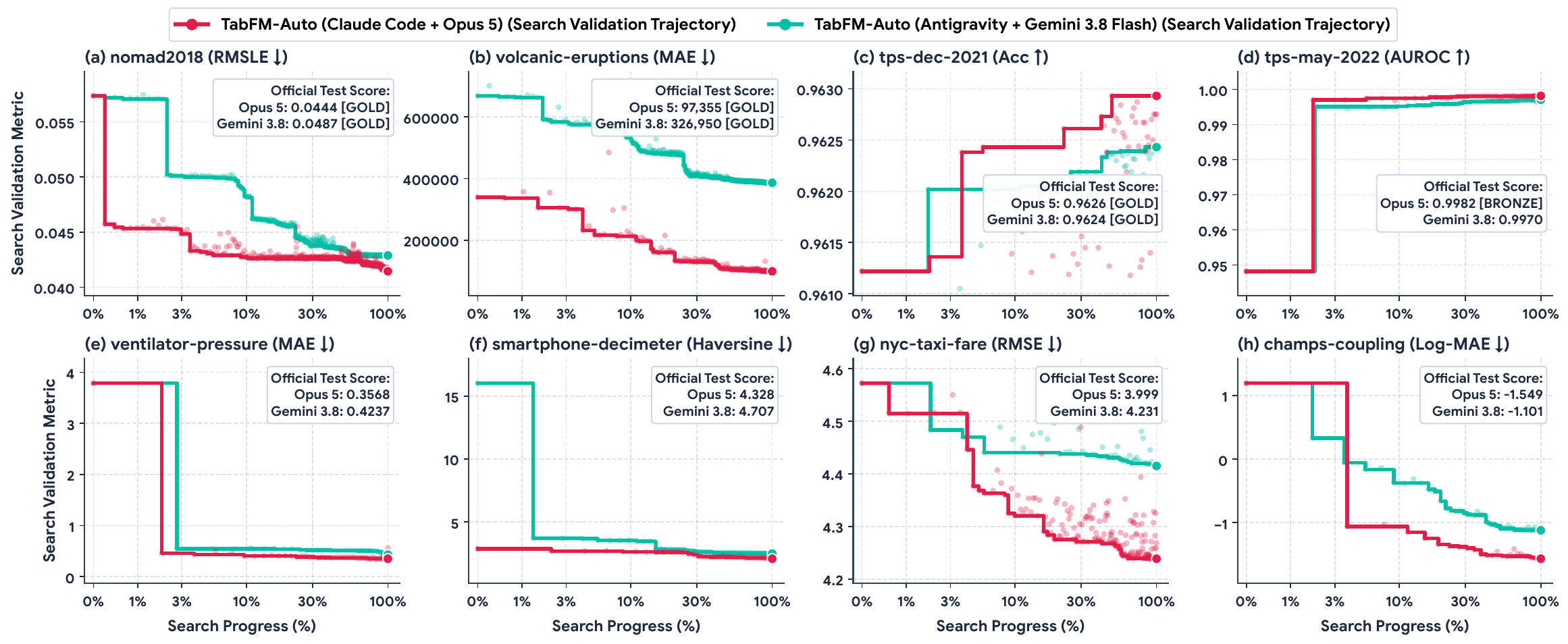}
    \caption{\textbf{Search progression on MLE-Bench-Tabular.} Best-so-far 3-fold cross-validation curves on the training split and final official test scores across the 8 Kaggle competitions (on \texttt{volcanic-eruptions} and \texttt{smartphone-decimeter}, where $P_0$ fails with zero features, $0\%$ plots each agent's step-2 initial multi-file script).}
    \label{fig:mlebench_curves}
\end{figure}

Figure~\ref{fig:mlebench_curves} plots the 3-fold cross-validation curves on the training split and the final official test scores across the 8 Kaggle competitions of MLE-Bench-Tabular. Table~\ref{tab:mlebench_results} summarizes the data formats, evaluation metrics, baseline scores, final test results, and main features engineered by \tabfmauto{} (\texttt{Claude Code} with \texttt{Opus 5}) and \tabfmauto{} (\texttt{Antigravity} with \texttt{Gemini 3.8 Flash}). Figure~\ref{fig:mlebench_elo} reports overall pairwise Elo ratings computed across all graded submissions against the 12 external MLE agents with complete 8-competition submissions from the official MLE-Bench leaderboard (\texttt{PiEvolve 24h} only has public submissions on 5 of the 8 competitions and, if rated on those 5, scores 1668 Elo, still behind both \tabfmauto{} configurations). \tabfmauto{} (\texttt{Claude Code} with \texttt{Opus 5}) ranks \#1 with 1827 Elo ($86.7\%$ win rate across all graded matches) and \tabfmauto{} (\texttt{Antigravity} with \texttt{Gemini 3.8 Flash}) ranks \#2 with 1734 Elo ($80.0\%$ win rate).

\paragraph{Geophysics and Multi-Sensor Waveforms (\texttt{predict-volcanic-eruptions}).}
The main training table in \texttt{predict-volcanic-eruptions} contains only segment IDs and time-to-eruption targets, while the sensor readings are stored in thousands of external 10-minute, 100\,Hz ten-channel seismic CSV files ($60{,}001$ rows per file). The starting identity pipeline $P_0$ therefore fails on step~1 with zero feature columns. The coding agent then inspects the error and writes an initial summary-statistics script on step~2 (scoring a validation mean absolute error, or MAE, of $667{,}762$ for \texttt{Gemini 3.8 Flash} and $339{,}254$ for \texttt{Opus 5}). It then refines $\Phi_{\text{feat}}$ into a parallelized signal-processing pipeline that computes multi-band Fast Fourier Transform (FFT) spectral energy across volcanic tremor bands ($0.5\text{--}20$\,Hz), Short-Term Average to Long-Term Average (STA/LTA) trigger ratios across sub-windows, rolling peak-to-peak envelopes, and inter-sensor cross-correlations. Passing this feature table to frozen \tabfm{} lowers official test MAE to $326{,}950$ (\texttt{Gemini 3.8 Flash}) and $97{,}355$ (\texttt{Opus 5}), reducing error by $9.4\times$ over the strongest external agent (\texttt{PiEvolve} at $0.92\text{M}$, \texttt{MLEvolve} at $1.24\text{M}$) on \texttt{MLE-Bench}'s official local split and earning a Gold Medal.

\paragraph{Materials Science and 3D Crystal Lattices (\texttt{nomad2018-predict-transparent-conductors}).}
The task is to predict formation energy and bandgap energy (scored by mean column-wise root mean squared logarithmic error, or RMSLE) of $(\text{Al}_x\text{Ga}_y\text{In}_z)_2\text{O}_3$ transparent conductors. The competition provides lattice vectors $(a, b, c, \alpha, \beta, \gamma)$ in the main table alongside external \texttt{geometry.xyz} files containing 3D Cartesian coordinates of the atoms in each unit cell. \tabfmauto{} parses the 3D crystal files to compute the parallelepiped unit-cell volume $V = abc\sqrt{1 + 2\cos\alpha\cos\beta\cos\gamma - \cos^2\alpha - \cos^2\beta - \cos^2\gamma}$, atomic packing density, reciprocal lattice constants, and composition-weighted Pauling electronegativity and ionic radius variance across the $\text{Al}/\text{Ga}/\text{In}$ cations. These physical features reduce official test RMSLE from $0.05737$ (\tabfm{}) to $0.04869$ (\texttt{Gemini 3.8 Flash}) and $0.04440$ (\texttt{Opus 5}), earning a Kaggle Gold Medal and ranking \#1 among all evaluated agents.

\paragraph{Quantum Chemistry and 3D Molecular Coupling (\texttt{champs-scalar-coupling}).}
Predicting nuclear magnetic resonance (NMR) spin-spin scalar coupling constants ($J$-coupling, evaluated by mean Log-MAE across 8 coupling types such as \texttt{1JHC} and \texttt{3JHH}) requires joining the bond-pair table with 3D molecular coordinates in \texttt{structures.csv}. Without 3D geometry, \tabfm{} scores $+1.19025$ Log-MAE. \tabfmauto{} writes a geometry pipeline in $\Phi_{\text{feat}}$ that computes 3D inter-atomic distances $r_{ij}$ and inverse power laws ($r_{ij}^{-1}, r_{ij}^{-2}, r_{ij}^{-3}$), covalent bond-path angles, and four-atom dihedral torsion angles $\phi$ from the Karplus relation ($\cos\phi$ and $\cos^2\phi$) for vicinal \texttt{3J} couplings, while routing each coupling type through stratified context windows in $\mathcal{S}_{\text{ctx}}$. This lowers official test Log-MAE to $-1.10133$ (\texttt{Gemini 3.8 Flash}) and $-1.54947$ (\texttt{Opus 5}).

\paragraph{Respiratory Control Dynamics (\texttt{ventilator-pressure-prediction}).}
This competition provides breath-by-breath time series (80 time steps per breath) of inspiratory solenoid valve opening percentages $u_{\text{in}}(t) \in [0, 100]$, expiratory valve states $u_{\text{out}}(t)$, and lung constants ($R$ resistance and $C$ compliance) to predict airway pressure. Treating time steps as independent rows in \tabfm{} gives $3.78450$ MAE because it ignores the accumulated air in the lung. Inspired by the single-compartment respiratory mechanics relation $P(t) = P_0 + R q(t) + \frac{1}{C}\int_0^t q(\tau)\,d\tau$, \tabfmauto{} uses $u_{\text{in}}(t)$ as a flow proxy to compute within-breath cumulative integrals $\int_0^t u_{\text{in}}(\tau)\,d\tau$, first and second finite differences ($\Delta u_{\text{in}}, \Delta^2 u_{\text{in}}$), decayed flow histories, and proxy interaction terms ($R \times u_{\text{in}}$ and $\frac{1}{C}\int_0^t u_{\text{in}}(\tau)\,d\tau$). These state features reduce official test MAE to $0.42369$ (\texttt{Gemini 3.8 Flash}) and $0.35684$ (\texttt{Opus 5}, a $90.6\%$ MAE reduction).

\paragraph{GNSS Satellite Navigation (\texttt{smartphone-decimeter-2022}).}
Predicting phone positions (scored by the mean of the 50th and 95th percentile horizontal Haversine distance errors in meters across trips) from raw Android Global Navigation Satellite System (GNSS) receiver logs requires joining drive traces with multi-gigabyte satellite pseudorange and Inertial Measurement Unit (IMU) tables (with ground-truth NMEA logs excluded). After the zero-feature identity pipeline $P_0$ fails on step~1, the coding agent extracts an initial weighted least squares (WLS) baseline from the receiver logs on step~2 ($16.010$\,m validation error for \texttt{Gemini 3.8 Flash} and $2.850$\,m for \texttt{Opus 5}). It then computes elevation-weighted carrier-to-noise density ($C/N_0$) averages, pseudorange residual dispersion, and forward-backward Kalman velocity-smoothed positions across consecutive epochs, reducing official test Haversine error to $4.7068$\,m (\texttt{Gemini 3.8 Flash}) and $4.3284$\,m (\texttt{Opus 5}, ranking \#1 overall and $18.9\%$ below \texttt{Famou-Agent 2.0} at $5.334$\,m).

\input{tables/mlebench_results}

\paragraph{Cartography, Manufacturing Telemetry, and Spatial Econometrics (\texttt{tps-dec-2021}, \texttt{tps-may-2022}, and \texttt{nyc-taxi-fare}).}
On the remaining three large-scale tabular competitions, \tabfmauto{} extracts geometric and structural features from the schema. All scores below are for \texttt{Opus 5}. On \texttt{tabular-playground-series-dec-2021}, computing Euclidean distance to hydrology $\sqrt{d_{\text{horiz}}^2 + d_{\text{vert}}^2}$, relative elevation above hydrology, and compass aspect ($\sin\theta, \cos\theta$) raises official test accuracy to $0.96264$ (Gold Medal). On \texttt{tabular-playground-series-may-2022}, splitting the 10-character manufacturing string \texttt{f\_27} into 10 positional ASCII ordinal columns, unique character counts, and continuous interaction terms reduces test $1-\mathrm{AUROC}$ error from $0.05179$ to $0.00179$ (Bronze Medal). On \texttt{new-york-city-taxi-fare-prediction}, computing Haversine and $29^\circ$-rotated Manhattan street-grid distances along with bounding-box flags for JFK, LaGuardia, and Newark airport flat-rate zones reduces test RMSE by $12.5\%$ from $4.57207$ to $3.99861$.

\section{Per-Dataset Official Test Scores on TabArena}
\label{sec:app_per_dataset}

Table~\ref{tab:all_51_datasets} reports the per-dataset official test scores across all 51 TabArena datasets, grouped into the 13 regression datasets evaluated by RMSE (Part~I), the 30 binary classification datasets evaluated by $1-\mathrm{AUROC}$, for which lower is better (Part~II), and the 8 multiclass classification datasets evaluated by multiclass log-loss (Part~III).

On Part~I (Regression), \tabfmauto{} improves official test RMSE over \tabfm{} on all 13 datasets, achieving a suite G-Mean RMSE reduction of $+3.11\%$ (\texttt{Opus 5}) and $+3.15\%$ (\texttt{Gemini 3.8 Flash}) compared to $+1.35\%$ for \tabfmp{}. The largest relative RMSE reductions occur on physical and engineering tasks where column names indicate domain formulas, including \texttt{airfoil\_self\_noise} ($+17.3\%$, \texttt{Gemini 3.8 Flash}), \texttt{physiochemical\_protein} ($+3.9\%$, \texttt{Opus 5}), \texttt{concrete\_strength} ($+3.8\%$, \texttt{Opus 5}), and \texttt{miami\_housing} ($+3.1\%$, \texttt{Opus 5}).

\input{tables/table_all_51_datasets}

On Part~II (Binary Classification), \tabfmauto{} achieves a suite G-Mean reduction in test classification error ($1 - \mathrm{AUROC}$) of $+3.51\%$ (\texttt{Opus 5}) and $+5.17\%$ (\texttt{Gemini 3.8 Flash}) vs.\ $+0.81\%$ for \tabfmp{}. The largest reductions in $1-\mathrm{AUROC}$ occur on \texttt{Amazon\_employee\_access} ($0.1416 \to 0.1039$, \texttt{Opus 5}), \texttt{churn} ($0.0681 \to 0.0491$, \texttt{Gemini 3.8 Flash}), \texttt{qsar-biodeg} ($0.0580 \to 0.0413$, \texttt{Gemini 3.8 Flash}), and \texttt{Marketing\_Campaign} ($0.0732 \to 0.0581$, \texttt{Gemini 3.8 Flash}). Error increases are small (at most $+0.0034$ in absolute $1-\mathrm{AUROC}$). The largest relative increases occur on nearly saturated datasets where \tabfm{} already attains $1-\mathrm{AUROC}$ around or below $0.01$ (\texttt{customer\_satisfaction\_in\_airline} at $0.0038$, \texttt{polish\_bankruptcy} at $0.0049$, \texttt{NATICUSdroid} at $0.0112$), and the largest absolute increases occur on small, noisy credit tables (\texttt{credit-g} with $N=1{,}000$, and \texttt{Is-this-a-good-customer}), where validation gains on fold~0 do not fully transfer across small evaluation folds. All other increases are at most $+0.0005$.

On Part~III (Multiclass Classification), \tabfmauto{} achieves its largest suite-level G-Mean improvement, reducing official test log-loss by $+8.39\%$ (\texttt{Opus 5}) and $+7.64\%$ (\texttt{Gemini 3.8 Flash}) compared to $+1.50\%$ for \tabfmp{}, improving 7 out of 8 datasets. Combining domain feature engineering in $\Phi_{\text{feat}}$ with multiclass temperature scaling and log-odds prior alignment in $\Psi_{\text{post}}$ produces the largest test log-loss reductions on \texttt{splice} ($+22.2\%$, \texttt{Opus 5}), \texttt{SDSS17} ($+18.8\%$, \texttt{Gemini 3.8 Flash}), \texttt{anneal} ($+17.3\%$, \texttt{Opus 5}), and \texttt{website\_phishing} ($+3.8\%$, \texttt{Gemini 3.8 Flash}). On all 51 datasets, \tabfmauto{} reduces overall G-Mean test error by $+4.19\%$ (\texttt{Opus 5}) and $+5.05\%$ (\texttt{Gemini 3.8 Flash}), roughly four to five times the $+1.06\%$ error reduction of feature engineering that ignores column meanings (cross and SVD features) and test-time ensembling in \tabfmp{}.

%% file: tables/leaderboard_overall.tex
% Generated by tabfm_auto_paper/scripts/generate_tabfm_auto_winrates_and_tables.py
% Matches exact 6-column aesthetic of tabfm_tech_report/tables/
% Requires in preamble:
%   \definecolor{tabfmautored}{HTML}{D93025}
%   \definecolor{tabfmblue}{HTML}{3186FF}
%   \newcommand{\best}[1]{\textbf{#1}}
% 95% bootstrap percentile interval of each Elo rating, relative to the median,
% rendered as a small coloured super/subscript pair.
\providecolor{elosdcol}{HTML}{5B7C99}
\providecommand{\eloci}[2]{\textcolor{elosdcol}{\ensuremath{{}^{\,+#2}_{\,-#1}}}}
\begin{table}[t]
\centering
\caption{\textbf{TabArena overall leaderboard across all 51 datasets.} Super/subscripts give the 95\% bootstrap percentile interval of each Elo rating relative to its median over 100 resampling rounds, and are asymmetric because the Bradley--Terry bootstrap distribution is right-skewed.}
\label{tab:overall}
\small
\renewcommand{\arraystretch}{1.45}
\resizebox{\linewidth}{!}{%
\begin{tabular}{clcccc}
\toprule
\textbf{\#} & \textbf{Method} & \textbf{Elo} $\uparrow$ & \textbf{Wins} $\uparrow$ & \textbf{Improv.\ (\%)} $\downarrow$ & \textbf{G-Mean} $\downarrow$\\
\midrule
\rowcolor{tabfmautored!12}
1 & TabFM-Auto (Codex, Opus 5) & \best{2013.0}\eloci{103.2}{123.7} & 25.00 & 1.42\% & 0.3400\\
\rowcolor{tabfmautored!12}
2 & TabFM-Auto (Claude Code, Opus 5) & 1993.6\eloci{124.3}{126.8} & 25.47 & \best{1.26\%} & 0.3402\\
\rowcolor{tabfmautored!12}
3 & TabFM-Auto (Antigravity, Gemini 3.8 Flash) & 1979.6\eloci{95.7}{111.3} & \best{26.12} & 1.82\% & \best{0.3371}\\
\rowcolor{tabfmautored!12}
4 & TabFM-Auto (Claude Code, Gemini 3.8 Flash) & 1957.8\eloci{103.7}{106.1} & 22.67 & 1.88\% & 0.3446\\
\rowcolor{tabfmautored!12}
5 & TabFM-Auto (Codex, Gemini 3.8 Flash) & 1940.1\eloci{133.9}{120.1} & 25.06 & 1.81\% & 0.3435\\[3pt]
\addlinespace[2pt]
6 & TabFM+ & 1856.0\eloci{102.7}{118.4} & 17.58 & 2.20\% & 0.3513\\
7 & TabFM & 1785.3\eloci{125.9}{114.9} & 9.74 & 3.20\% & 0.3550\\
8 & EXAONE-Tabular & 1764.6\eloci{72.4}{90.9} & 5.56 & 6.05\% & 0.3703\\
9 & AutoGluon 1.5 (extreme) & 1668.4\eloci{79.2}{80.6} & 2.11 & 6.79\% & 0.3718\\
10 & TabPFN-3 & 1660.1\eloci{69.6}{84.0} & 1.61 & 8.14\% & 0.3811\\
11 & AutoGluon 1.4 (4h) & 1612.3\eloci{61.3}{73.0} & 0.58 & 9.50\% & 0.3913\\
12 & TabPFN-2.6 & 1604.8\eloci{73.9}{71.3} & 0.28 & 9.69\% & 0.3896\\
13 & TabICLv2 & 1590.1\eloci{87.9}{114.1} & 1.92 & 9.04\% & 0.3856\\
14 & RealTabPFN-2.5 & 1521.9\eloci{65.7}{92.8} & 0.10 & 10.38\% & 0.3918\\
15 & AutoGluon 1.3 (4h) & 1490.3\eloci{79.6}{67.1} & 0.18 & 12.15\% & 0.4115\\
16 & TabDPT-Turbo & 1455.8\eloci{71.3}{70.0} & 0.67 & 13.11\% & 0.4127\\
\bottomrule
\end{tabular}%
}
\end{table}

%% file: tables/leaderboard_fold0.tex
% 95% bootstrap percentile interval of each Elo rating, relative to the median,
% rendered as a small coloured super/subscript pair.
\providecolor{elosdcol}{HTML}{5B7C99}
\providecommand{\eloci}[2]{\textcolor{elosdcol}{\ensuremath{{}^{\,+#2}_{\,-#1}}}}
\begin{table}[t]
\centering
\caption{\textbf{TabArena Fold-0 held-out test Elo ratings across Overall (51 datasets), Classification (38 datasets), and Regression (13 datasets).} Every method is scored on the official held-out test split of Fold~0, which shares no rows with Fold~0's training split used for 3-fold cross-validation to select $P^*$. As in Table~\ref{tab:overall}, each \tabfmauto{} configuration is rated individually against the 66-method TabArena pool using bootstrap Bradley--Terry estimation anchored to \texttt{RandomForest (default)} $= 1000$. Super/subscripts give the 95\% bootstrap percentile interval of each rating relative to its median.}
\label{tab:fold0_elo}
\resizebox{\textwidth}{!}{%
\setlength{\tabcolsep}{7.5pt}
\renewcommand{\arraystretch}{1.45}
\begin{tabular}{clccc}
\toprule
\textbf{\#} & \textbf{Method} & \textbf{Overall Elo (51)} $\uparrow$ & \textbf{Classification Elo (38)} $\uparrow$ & \textbf{Regression Elo (13)} $\uparrow$ \\
\midrule
\rowcolor{tabfmautored!12}
1 & \tabfmauto{} (\texttt{Codex}, \texttt{Opus 5}) & \best{1950.7}\eloci{169.3}{190.3} & \best{1877.5}\eloci{183.6}{197.0} & 2616.3\eloci{188.8}{355.1} \\
\rowcolor{tabfmautored!12}
2 & \tabfmauto{} (\texttt{Claude Code}, \texttt{Gemini 3.8 Flash}) & 1862.8\eloci{163.7}{181.0} & 1805.0\eloci{187.6}{186.1} & 2336.1\eloci{318.9}{505.0} \\
3 & \tabfmp{} & 1853.2\eloci{177.3}{194.2} & 1782.8\eloci{187.4}{204.9} & 2409.9\eloci{173.8}{294.5} \\
\rowcolor{tabfmautored!12}
4 & \tabfmauto{} (\texttt{Claude Code}, \texttt{Opus 5}) & 1843.3\eloci{172.0}{202.8} & 1763.6\eloci{184.8}{203.7} & 2562.2\eloci{217.6}{422.8} \\
\rowcolor{tabfmautored!12}
5 & \tabfmauto{} (\texttt{Antigravity}, \texttt{Gemini 3.8 Flash}) & 1791.9\eloci{155.4}{181.7} & 1710.5\eloci{154.3}{201.8} & 2532.4\eloci{206.1}{319.5} \\
6 & \tabfm{} & 1790.2\eloci{168.5}{181.4} & 1733.0\eloci{190.7}{201.1} & 2227.2\eloci{144.8}{262.9} \\
\rowcolor{tabfmautored!12}
7 & \tabfmauto{} (\texttt{Codex}, \texttt{Gemini 3.8 Flash}) & 1785.6\eloci{187.4}{239.5} & 1703.2\eloci{193.9}{224.5} & \best{2645.6}\eloci{184.3}{353.7} \\
8 & \texttt{EXAONE-Tabular} & 1772.2\eloci{128.9}{133.7} & 1740.4\eloci{140.8}{169.0} & 2032.4\eloci{180.0}{211.4} \\
9 & \texttt{AutoGluon 1.5} (extreme, 4h) & 1661.0\eloci{136.4}{136.1} & 1612.8\eloci{160.1}{148.7} & 2019.1\eloci{81.3}{126.6} \\
10 & TabPFN-3 & 1653.7\eloci{112.5}{125.6} & 1608.0\eloci{126.8}{135.8} & 1971.0\eloci{186.6}{294.0} \\
11 & TabPFN-2.6 & 1626.9\eloci{112.8}{110.6} & 1582.3\eloci{118.0}{125.9} & 1916.7\eloci{121.8}{195.7} \\
12 & AutoGluon 1.4 (4h) & 1596.2\eloci{123.7}{121.0} & 1568.9\eloci{151.1}{143.2} & 1800.7\eloci{99.7}{146.1} \\
13 & TabICLv2 & 1589.9\eloci{122.9}{132.3} & 1573.8\eloci{136.8}{159.0} & 1742.0\eloci{233.9}{355.4} \\
14 & RealTabPFN-2.5 & 1540.3\eloci{97.9}{104.1} & 1525.2\eloci{107.2}{121.8} & 1685.2\eloci{220.9}{248.6} \\
15 & AutoGluon 1.3 (4h) & 1505.9\eloci{93.4}{94.4} & 1473.6\eloci{106.2}{116.9} & 1720.6\eloci{114.2}{125.4} \\
16 & Mitra & 1452.5\eloci{119.0}{121.5} & 1439.5\eloci{139.3}{144.7} & 1586.7\eloci{171.4}{205.4} \\
17 & RealMLP (tuned + ensemble) & 1449.9\eloci{92.5}{88.8} & 1418.1\eloci{115.4}{103.0} & 1665.7\eloci{87.0}{114.1} \\
18 & TabDPT-Turbo & 1418.0\eloci{83.6}{94.7} & 1373.5\eloci{102.8}{110.9} & 1697.1\eloci{108.2}{146.7} \\
19 & LightGBM (tuned + ensemble) & 1392.8\eloci{79.8}{83.7} & 1366.9\eloci{93.2}{107.3} & 1566.8\eloci{83.2}{96.0} \\
20 & TabM (tuned + ensemble) & 1387.6\eloci{103.0}{103.7} & 1372.7\eloci{129.6}{135.5} & 1502.7\eloci{118.9}{145.5} \\
21 & CatBoost (tuned + ensemble) & 1379.0\eloci{95.7}{89.9} & 1356.3\eloci{116.8}{109.7} & 1523.6\eloci{89.1}{124.7} \\
\bottomrule
\end{tabular}%
}
\end{table}

%% file: tables/feature_engineering_taxonomy.tex
\begin{table}[t]
\centering
\caption{\textbf{Taxonomy of discovered pipelines on TabArena.} Each final pipeline $P^*$ from \texttt{Claude Code} with \texttt{Opus 5} and \texttt{Antigravity} with \texttt{Gemini 3.8 Flash} across the 51 datasets is classified into one of three mutually exclusive categories. \textit{Configuration} lists each agent setup as well as \textit{Combined Best} (selecting the pipeline with the higher 3-fold cross-validation score $\mathcal{U}_{\text{val}}$ between the two configurations per dataset). \textit{Datasets} reports the fraction of the 51 TabArena datasets assigned to each category, and \textit{Mean} and \textit{Median} report the mean and median relative official test error reduction ($\Delta\%$) over \tabfm{} across datasets in that category (RMSE for regression, $1-\mathrm{AUROC}$ for binary classification, and log-loss for multiclass classification).}
\label{tab:feature_engineering_taxonomy}
\setlength{\tabcolsep}{4.8pt}
\renewcommand{\arraystretch}{1.15}
\resizebox{\textwidth}{!}{%
\begin{tabular}{p{2.6cm}p{10.2cm}lccc}
\toprule
\textbf{Category} & \textbf{Representative Features Built in Code} & \textbf{Configuration} & \textbf{Datasets} & \textbf{Mean ($\Delta\%$)} & \textbf{Median ($\Delta\%$)} \\
\midrule
\multirow{3}{2.6cm}{\textbf{I: Domain-Knowledge Features}}
 & \textbf{Physics \& Engineering}: Strouhal $St=f\delta^*/U_\infty$, Helmholtz $He$ (\texttt{airfoil}); water-to-binder ratio (\texttt{concrete}) & Claude Code, Opus 5 & 17 / 51 & \textbf{+7.25\%} & \textbf{+3.01\%} \\
 & \textbf{Clinical Medicine}: ICD-9 organ chapters (\texttt{Diabetes130US}); $\text{MAP}=(2\text{DBP}+\text{SBP})/3$, Shock Index (\texttt{MIC}) & Antigravity, Gemini 3.8 Flash & 17 / 51 & \textbf{+8.16\%} & \textbf{+2.74\%} \\
 & \textbf{Science \& Genomics}: Coil $\texttt{width}=0\to\text{NaN}$ (\texttt{anneal}); SDSS color indices (\texttt{SDSS17}); DNA splice motifs (\texttt{splice}) & \cellcolor{tabfmautored!12}\textit{Combined Best} & \cellcolor{tabfmautored!12}\textit{17 / 51} & \cellcolor{tabfmautored!12}\textit{\textbf{+8.85\%}} & \cellcolor{tabfmautored!12}\textit{\textbf{+3.01\%}} \\
\midrule
\multirow{3}{2.6cm}{\textbf{II: Statistical \& Structural Features}}
 & \textbf{Relational Graph Degrees}: Marginal entity counts, pairwise co-occurrences, distinct-$B$-per-$A$ (\texttt{Amazon\_employee}) & Claude Code, Opus 5 & 34 / 51 & +2.31\% & +0.90\% \\
 & \textbf{Frequency \& Missingness}: Level counts, cross-fitted target means, missingness-aware column ranking (\texttt{kddcup09\_appetency}) & Antigravity, Gemini 3.8 Flash & 29 / 51 & +2.99\% & +1.32\% \\
 & \textbf{Wide-Table Compression}: Collinearity pruning and truncated SVD projections (\texttt{Bioresponse}, \texttt{hiva\_agnostic}) & \cellcolor{tabfmblue!12}\textit{Combined Best} & \cellcolor{tabfmblue!12}\textit{34 / 51} & \cellcolor{tabfmblue!12}\textit{+3.87\%} & \cellcolor{tabfmblue!12}\textit{+1.71\%} \\
\midrule
\multirow{2}{2.6cm}{\textbf{III: Context \& Calibration Only}}
 & \textbf{Context Selection}: Minority-class context oversampling (\texttt{sample()}) without modifying $\mathbf{X}$ & Claude Code, Opus 5 & 0 / 51 & --- & --- \\
 & \textbf{Prior Calibration}: Log-odds prior alignment (\texttt{postprocess()}) and temperature/vector scaling (\texttt{TABFM\_KWARGS}) & Antigravity, Gemini 3.8 Flash & 5 / 51 & +2.16\% & +0.19\% \\
\bottomrule
\end{tabular}%
}
\end{table}

%% file: tables/transfer_elo.tex
% Generated by tabfm_auto_paper/scripts/generate_transfer.py from data/elo_transfer.json.
\begin{table}[t]
\centering
\caption{\textbf{Final pipelines found for \tabfm{} also help other tabular foundation models.} TabArena Elo of each model with its default configuration and within the final per-dataset pipelines that \tabfmauto{} found for \tabfm{} with each coding agent and LLM. Each model is rated in a separate fit per configuration, and \emph{Gain} is measured against the default model rated in the same fit. \emph{Default} is the mean of the three default ratings. Bold marks the largest gain for each model.}
\label{tab:transfer}
\resizebox{\linewidth}{!}{%
\setlength{\tabcolsep}{8pt}
\renewcommand{\arraystretch}{1.15}
\begin{tabular}{llrrrrrr}
\toprule
\multirow{2}{*}{\textbf{Model}} & \multirow{2}{*}{\textbf{Pipelines}} & \multicolumn{2}{c}{\textbf{Overall (51)}} & \multicolumn{2}{c}{\textbf{Classification (38)}} & \multicolumn{2}{c}{\textbf{Regression (13)}} \\
\cmidrule(lr){3-4} \cmidrule(lr){5-6} \cmidrule(lr){7-8}
 & & Elo & Gain & Elo & Gain & Elo & Gain \\
\midrule
\multirow{4}{*}{\texttt{TabICLv2}} & Default & 1593.0 & & 1587.1 & & 1726.6 & \\
 & Codex, Opus 5 & 1732.8 & +139.4 & 1685.1 & +97.6 & 2088.9 & \best{+361.5} \\
 & Claude Code, Opus 5 & 1736.7 & \best{+143.3} & 1704.4 & \best{+116.8} & 2006.5 & +279.5 \\
 & Antigravity, Gemini 3.8 Flash & 1700.2 & +108.0 & 1678.2 & +92.1 & 1938.5 & +213.1 \\
\midrule
\multirow{4}{*}{\texttt{TabPFN-3}} & Default & 1666.0 & & 1643.8 & & 1869.3 & \\
 & Codex, Opus 5 & 1755.5 & +89.7 & 1705.1 & +61.8 & 2112.2 & +242.6 \\
 & Claude Code, Opus 5 & 1797.3 & \best{+130.8} & 1751.1 & \best{+106.5} & 2189.5 & \best{+318.5} \\
 & Antigravity, Gemini 3.8 Flash & 1777.0 & +111.2 & 1741.8 & +98.3 & 2078.5 & +211.3 \\
\midrule
\multirow{4}{*}{\texttt{EXAONE-Tabular}} & Default & 1773.6 & & 1770.0 & & 1971.6 & \\
 & Codex, Opus 5 & 1857.5 & +82.3 & 1839.6 & \best{+68.4} & 2088.0 & +118.6 \\
 & Claude Code, Opus 5 & 1862.0 & \best{+88.7} & 1817.2 & +46.4 & 2200.1 & \best{+225.8} \\
 & Antigravity, Gemini 3.8 Flash & 1841.2 & +68.8 & 1814.5 & +46.5 & 2083.8 & +112.7 \\
\bottomrule
\end{tabular}%
}
\end{table}

%% file: tables/mlebench_results.tex
\begin{table}[t]
\centering
\definecolor{goldbg}{HTML}{FEF3C7}\definecolor{goldfg}{HTML}{92400E}
\definecolor{silverbg}{HTML}{E2E8F0}\definecolor{silverfg}{HTML}{1E293B}
\definecolor{bronzebg}{HTML}{FFEDD5}\definecolor{bronzefg}{HTML}{9A3412}
\definecolor{greenbg}{HTML}{DCFCE7}\definecolor{greenfg}{HTML}{166534}
\caption{\textbf{End-to-end performance on 8 Kaggle MLE-Bench competitions.} Best score per task is \textbf{bolded}, second best is \underline{underlined}.}
\label{tab:mlebench_results}
\setlength{\tabcolsep}{4.5pt}
\renewcommand{\arraystretch}{1.15}
\resizebox{\textwidth}{!}{%
\begin{tabular}{llcccccp{5.4cm}}
\toprule
\textbf{Competition} & \textbf{Domain \& Data Format} & \textbf{Metric} & \textbf{Identity} & \textbf{AGY +} & \textbf{CC +} & \textbf{Result} & \textbf{Key Features Added in \texttt{engineer()}} \\
 & & & \textbf{TabFM} & \textbf{Gemini 3.8 Flash} & \textbf{Opus 5} & & \\
\midrule
\texttt{volcanic-eruptions} & Geophysics (Waveforms) & MAE $\downarrow$ & Fails (IDs)$^\dagger$ & \underline{326,950} & \textbf{97,355} & \cellcolor{goldbg}\textcolor{goldfg}{\textbf{GOLD}} & Multi-sensor waveform FFT bands, STA/LTA ratios, quantiles ($9.4\times <$ \texttt{PiEvolve}) \\
\texttt{nomad2018-conductors} & Materials (3D Lattice) & RMSLE $\downarrow$ & 0.05737 & \underline{0.04869} & \textbf{0.04440} & \cellcolor{goldbg}\textcolor{goldfg}{\textbf{GOLD}} & Reciprocal lattice unit-cell volume, cation electronegativity \& radius dispersion \\
\texttt{tps-dec-2021} & Cartography \& Hydrology & Acc $\uparrow$ & 0.96122 & \underline{0.96237} & \textbf{0.96264} & \cellcolor{goldbg}\textcolor{goldfg}{\textbf{GOLD}} & Euclidean hydrology distance $\sqrt{H^2+V^2}$, aspect compass decomposition \\
\texttt{tps-may-2022} & Manufacturing Telemetry & $1 - \mathrm{AUROC}$ $\downarrow$ & 0.05179 & \underline{0.00304} & \textbf{0.00179} & \cellcolor{bronzebg}\textcolor{bronzefg}{\textbf{BRONZE}} & Decomposing 10-char string \texttt{f\_27} into positional ordinal \& unique-char counts \\
\texttt{champs-coupling} & Quantum Chem (3D NMR) & Log-MAE $\downarrow$ & $+$1.19025 & \underline{$-$1.10133} & \textbf{$-$1.54947} & \cellcolor{greenbg}\textcolor{greenfg}{\textbf{$>$MEDIAN}} & 3D inter-atomic Euclidean distance $r_{ij}^{-3}$, Karplus dihedral $\cos^2\phi$ angles \\
\texttt{ventilator-pressure} & Respiratory Control & MAE $\downarrow$ & 3.78450 & \underline{0.42369} & \textbf{0.35684} & \cellcolor{greenbg}\textcolor{greenfg}{\textbf{$-$90.6\%}} & Respiratory mechanics proxy integrals ($\int u_{\text{in}}\,dt$, $\Delta u_{\text{in}}$, $R\times u_{\text{in}}$) \\
\texttt{smartphone-decimeter} & GNSS Navigation & Haversine $\downarrow$ & Fails (IDs)$^\dagger$ & \underline{4.7068} & \textbf{4.3284} & \cellcolor{greenbg}\textcolor{greenfg}{\textbf{$>$MEDIAN}} & Multi-table GNSS pseudorange WLS + IMU Kalman smoothing ($18.9\% <$ \texttt{Famou-Agent 2.0}) \\
\texttt{nyc-taxi-fare} & Spatial Econometrics & RMSE $\downarrow$ & 4.57207 & \underline{4.23091} & \textbf{3.99861} & \cellcolor{greenbg}\textcolor{greenfg}{\textbf{$-$12.5\%}} & Geodesic Haversine distance, Manhattan rotated grid, JFK/LGA/EWR polygons \\
\bottomrule
\multicolumn{8}{p{16.8cm}}{\footnotesize $^\dagger$Raw tabular identity ($P_0$) fails on \texttt{volcanic-eruptions} and \texttt{smartphone-decimeter} because the main table contains only recording IDs; on step~2 the coding agent inspects the error and writes an initial multi-file script itself (scoring $667{,}762$ and $339{,}254$ validation MAE on \texttt{volcanic-eruptions}, and $16.010$\,m and $2.850$\,m validation Haversine error on \texttt{smartphone-decimeter}, for \texttt{Gemini 3.8 Flash} and \texttt{Opus 5}, respectively) before refining it.}
\end{tabular}%
}
\end{table}

%% file: tables/table_all_51_datasets.tex
% Auto-generated 51-Dataset Per-Dataset Results Table
\begin{table}[t]
\centering
\caption{\textbf{Per-Dataset Official Test Performance Across All 51 TabArena Benchmarks.} For each dataset, we report the number of official evaluation folds, \tabfm{}, and the final submitted pipelines discovered by \tabfmauto{} (\texttt{Claude Code} with \texttt{Opus 5}) and \tabfmauto{} (\texttt{Antigravity} with \texttt{Gemini 3.8 Flash}) across the Regression (RMSE $\downarrow$), Binary Classification ($1 - \mathrm{AUROC}$ $\downarrow$), and Multiclass Classification (Log-Loss $\downarrow$) suites. Parentheses report the relative test metric improvement ($\Delta\%$ RMSE reduction, $1 - \mathrm{AUROC}$ reduction, or Log-Loss reduction) vs.\ TabFM. Best score per row is \textbf{bolded}.}
\label{tab:all_51_datasets}
{\scriptsize
\renewcommand{\arraystretch}{0.88}
\setlength{\tabcolsep}{2.5pt}
\begin{tabular*}{\textwidth}{@{\extracolsep{\fill}}l c r r r@{}}
\toprule
\textbf{Dataset} & \textbf{Folds} & \textbf{\tabfm{}} & \textbf{TabFM-Auto (\texttt{Opus 5})} & \textbf{TabFM-Auto (\texttt{Gemini 3.8 Flash})} \\
\midrule
\multicolumn{5}{l}{\cellcolor{gray!12}\textbf{Part I: Regression Suite (13 Datasets - Metric: RMSE $\downarrow$)}} \\
\midrule
\texttt{airfoil\_self\_noise} & 30 & 1.0734 & 0.91650 \textcolor{teal}{(+14.6\%)} & \textbf{0.88780} \textcolor{teal}{(+17.3\%)} \\
\texttt{Used-Fiat-500} & 30 & 703.27 & \textbf{693.49} \textcolor{teal}{(+1.4\%)} & 696.52 \textcolor{teal}{(+1.0\%)} \\
\texttt{concrete\_strength} & 30 & 3.9666 & \textbf{3.8169} \textcolor{teal}{(+3.8\%)} & 3.8309 \textcolor{teal}{(+3.4\%)} \\
\texttt{diamonds} & 9 & 496.42 & \textbf{484.52} \textcolor{teal}{(+2.4\%)} & 484.54 \textcolor{teal}{(+2.4\%)} \\
\texttt{Food\_Delivery\_Time} & 9 & 7.3328 & 7.1736 \textcolor{teal}{(+2.2\%)} & \textbf{7.1724} \textcolor{teal}{(+2.2\%)} \\
\texttt{healthcare\_insurance} & 30 & 4,417.6 & \textbf{4,344.1} \textcolor{teal}{(+1.7\%)} & 4,392.3 \textcolor{teal}{(+0.6\%)} \\
\texttt{houses} & 9 & 0.18155 & 0.17706 \textcolor{teal}{(+2.5\%)} & \textbf{0.17696} \textcolor{teal}{(+2.5\%)} \\
\texttt{miami\_housing} & 9 & 74,069.5 & \textbf{71,802.2} \textcolor{teal}{(+3.1\%)} & 72,037.1 \textcolor{teal}{(+2.7\%)} \\
\texttt{physiochemical\_protein} & 9 & 2.8875 & \textbf{2.7751} \textcolor{teal}{(+3.9\%)} & 2.7838 \textcolor{teal}{(+3.6\%)} \\
\texttt{QSAR-TID-11} & 9 & 0.73934 & 0.72710 \textcolor{teal}{(+1.7\%)} & \textbf{0.72372} \textcolor{teal}{(+2.1\%)} \\
\texttt{QSAR\_fish\_toxicity} & 30 & 0.85345 & \textbf{0.84829} \textcolor{teal}{(+0.6\%)} & 0.85218 \textcolor{teal}{(+0.1\%)} \\
\texttt{superconductivity} & 9 & 8.9437 & \textbf{8.7946} \textcolor{teal}{(+1.7\%)} & 8.8023 \textcolor{teal}{(+1.6\%)} \\
\texttt{wine\_quality} & 9 & 0.58651 & \textbf{0.58551} \textcolor{teal}{(+0.2\%)} & 0.58622 \textcolor{gray}{(+0.0\%)} \\
\midrule
\textit{Suite G-Mean RMSE Reduction vs.\ TabFM} & --- & --- & \textbf{+3.11\%} & \textbf{+3.15\%} \\
\midrule
\multicolumn{5}{l}{\cellcolor{gray!12}\textbf{Part II: Binary Classification Suite (30 Datasets - Metric: $1 - \mathrm{AUROC}$ $\downarrow$)}} \\
\midrule
\texttt{Amazon\_employee\_access} & 9 & 0.1416 & \textbf{0.1039} \textcolor{teal}{(+26.6\%)} & 0.1063 \textcolor{teal}{(+25.0\%)} \\
\texttt{APSFailure} & 9 & 0.0056 & \textbf{0.0054} \textcolor{teal}{(+3.0\%)} & \textbf{0.0054} \textcolor{teal}{(+3.0\%)} \\
\texttt{bank-marketing} & 9 & 0.2310 & 0.2314 \textcolor{purple}{(-0.2\%)} & \textbf{0.2303} \textcolor{teal}{(+0.3\%)} \\
\texttt{Bank\_Customer\_Churn} & 9 & 0.1229 & 0.1228 \textcolor{teal}{(+0.1\%)} & \textbf{0.1225} \textcolor{teal}{(+0.3\%)} \\
\texttt{Bioresponse} & 9 & 0.1192 & 0.1171 \textcolor{teal}{(+1.8\%)} & \textbf{0.1168} \textcolor{teal}{(+2.0\%)} \\
\texttt{blood-transfusion-service-center} & 30 & 0.2441 & 0.2431 \textcolor{teal}{(+0.4\%)} & \textbf{0.2398} \textcolor{teal}{(+1.8\%)} \\
\texttt{churn} & 9 & 0.0681 & 0.0610 \textcolor{teal}{(+10.5\%)} & \textbf{0.0491} \textcolor{teal}{(+27.9\%)} \\
\texttt{coil2000\_insurance} & 9 & 0.2276 & \textbf{0.2127} \textcolor{teal}{(+6.5\%)} & 0.2225 \textcolor{teal}{(+2.2\%)} \\
\texttt{credit-g} & 30 & 0.1944 & \textbf{0.1940} \textcolor{teal}{(+0.2\%)} & 0.1978 \textcolor{purple}{(-1.8\%)} \\
\texttt{credit\_card\_default} & 9 & \textbf{0.2068} & 0.2073 \textcolor{purple}{(-0.2\%)} & 0.2073 \textcolor{purple}{(-0.2\%)} \\
\texttt{customer\_satisfaction\_in\_airline} & 9 & \textbf{0.0038} & 0.0038 \textcolor{purple}{(-0.9\%)} & 0.0038 \textcolor{purple}{(-1.6\%)} \\
\texttt{diabetes} & 30 & 0.1580 & \textbf{0.1471} \textcolor{teal}{(+6.9\%)} & 0.1550 \textcolor{teal}{(+1.9\%)} \\
\texttt{Diabetes130US} & 9 & 0.3292 & \textbf{0.3193} \textcolor{teal}{(+3.0\%)} & 0.3207 \textcolor{teal}{(+2.6\%)} \\
\texttt{E-CommereShippingData} & 9 & 0.2592 & 0.2508 \textcolor{teal}{(+3.2\%)} & \textbf{0.2412} \textcolor{teal}{(+6.9\%)} \\
\texttt{Fitness\_Club} & 30 & 0.1789 & 0.1787 \textcolor{teal}{(+0.1\%)} & \textbf{0.1785} \textcolor{teal}{(+0.2\%)} \\
\texttt{GiveMeSomeCredit} & 9 & 0.1321 & 0.1305 \textcolor{teal}{(+1.2\%)} & \textbf{0.1303} \textcolor{teal}{(+1.3\%)} \\
\texttt{hazelnut-spread-contaminant} & 30 & 0.0023 & \textbf{0.0021} \textcolor{teal}{(+6.7\%)} & 0.0021 \textcolor{teal}{(+6.4\%)} \\
\texttt{heloc} & 9 & 0.1976 & 0.1973 \textcolor{teal}{(+0.2\%)} & \textbf{0.1971} \textcolor{teal}{(+0.3\%)} \\
\texttt{HR\_Analytics\_Job\_Change} & 9 & 0.1935 & \textbf{0.1933} \textcolor{teal}{(+0.1\%)} & 0.1940 \textcolor{purple}{(-0.3\%)} \\
\texttt{in\_vehicle\_coupon} & 9 & 0.1513 & 0.1405 \textcolor{teal}{(+7.2\%)} & \textbf{0.1402} \textcolor{teal}{(+7.3\%)} \\
\texttt{Is-this-a-good-customer} & 30 & 0.2466 & \textbf{0.2432} \textcolor{teal}{(+1.4\%)} & 0.2499 \textcolor{purple}{(-1.3\%)} \\
\texttt{jm1} & 9 & 0.2109 & 0.2114 \textcolor{purple}{(-0.3\%)} & \textbf{0.2094} \textcolor{teal}{(+0.7\%)} \\
\texttt{kddcup09\_appetency} & 9 & 0.1587 & \textbf{0.1505} \textcolor{teal}{(+5.2\%)} & 0.1515 \textcolor{teal}{(+4.5\%)} \\
\texttt{Marketing\_Campaign} & 30 & 0.0732 & 0.0616 \textcolor{teal}{(+15.8\%)} & \textbf{0.0581} \textcolor{teal}{(+20.7\%)} \\
\texttt{NATICUSdroid} & 9 & \textbf{0.0112} & 0.0115 \textcolor{purple}{(-2.6\%)} & 0.0122 \textcolor{purple}{(-9.3\%)} \\
\texttt{online\_shoppers\_intention} & 9 & \textbf{0.0601} & 0.0602 \textcolor{purple}{(-0.3\%)} & 0.0605 \textcolor{purple}{(-0.7\%)} \\
\texttt{polish\_bankruptcy} & 9 & \textbf{0.0049} & 0.0052 \textcolor{purple}{(-6.5\%)} & 0.0054 \textcolor{purple}{(-11.5\%)} \\
\texttt{qsar-biodeg} & 30 & 0.0580 & 0.0582 \textcolor{purple}{(-0.3\%)} & \textbf{0.0413} \textcolor{teal}{(+28.7\%)} \\
\texttt{seismic-bumps} & 9 & 0.2047 & \textbf{0.1987} \textcolor{teal}{(+2.9\%)} & 0.2022 \textcolor{teal}{(+1.2\%)} \\
\texttt{taiwanese\_bankruptcy} & 9 & 0.0494 & 0.0459 \textcolor{teal}{(+7.1\%)} & \textbf{0.0395} \textcolor{teal}{(+20.0\%)} \\
\midrule
\textit{Suite G-Mean ($1-\mathrm{AUROC}$) Reduction vs.\ TabFM} & --- & --- & \textbf{+3.51\%} & \textbf{+5.17\%} \\
\midrule
\multicolumn{5}{l}{\cellcolor{gray!12}\textbf{Part III: Multiclass Classification Suite (8 Datasets - Metric: Log-Loss $\downarrow$)}} \\
\midrule
\texttt{anneal} & 30 & 0.0125 & \textbf{0.0103} \textcolor{teal}{(+17.3\%)} & 0.0114 \textcolor{teal}{(+8.9\%)} \\
\texttt{hiva\_agnostic} & 9 & 0.1787 & 0.1779 \textcolor{teal}{(+0.4\%)} & \textbf{0.1742} \textcolor{teal}{(+2.5\%)} \\
\texttt{maternal\_health\_risk} & 30 & 0.3706 & 0.3648 \textcolor{teal}{(+1.6\%)} & \textbf{0.3610} \textcolor{teal}{(+2.6\%)} \\
\texttt{MIC} & 30 & 0.4282 & \textbf{0.4178} \textcolor{teal}{(+2.4\%)} & 0.4199 \textcolor{teal}{(+1.9\%)} \\
\texttt{SDSS17} & 9 & 0.0693 & 0.0564 \textcolor{teal}{(+18.6\%)} & \textbf{0.0563} \textcolor{teal}{(+18.8\%)} \\
\texttt{splice} & 9 & 0.0960 & \textbf{0.0747} \textcolor{teal}{(+22.2\%)} & 0.0757 \textcolor{teal}{(+21.1\%)} \\
\texttt{students\_dropout\_and\_academic\_success} & 9 & \textbf{0.5068} & 0.5111 \textcolor{purple}{(-0.8\%)} & 0.5132 \textcolor{purple}{(-1.2\%)} \\
\texttt{website\_phishing} & 30 & 0.2104 & 0.2067 \textcolor{teal}{(+1.8\%)} & \textbf{0.2024} \textcolor{teal}{(+3.8\%)} \\
\midrule
\textit{Suite G-Mean Log-Loss Reduction vs.\ TabFM} & --- & --- & \textbf{+8.39\%} & \textbf{+7.64\%} \\
\midrule
\rowcolor{teal!10} \textbf{Overall G-Mean Error Reduction (All 51 Datasets)} & --- & \textbf{0.00\%} & \textbf{+4.19\%} & \textbf{+5.05\%} \\
\bottomrule
\end{tabular*}%
}
\end{table}